\documentclass[10pt,twocolumn,letterpaper]{article}
\usepackage[pagenumbers]{cvpr}
\usepackage[T1]{fontenc}
\definecolor{cvprblue}{rgb}{0.21,0.49,0.74}

\usepackage[pagebackref,breaklinks,colorlinks,allcolors=cvprblue]{hyperref}
\usepackage[capitalize]{cleveref}
\usepackage{placeins}
\usepackage{wrapfig}
\usepackage{xspace}
\hypersetup{
  pdftitle={Kiln: Asynchronous Flow for Artist Mesh Generation, Editing, and Refinement},
  pdfauthor={Anonymous},
  pdfsubject={Anonymous manuscript draft for CVPR 2027}
}
\graphicspath{{figures/}}
\newcommand{\name}{Kiln\xspace}

\providecommand{\KC}[1]{{\color{blue}#1}}
\AtBeginDocument{
  \crefname{figure}{Fig.}{Fig.}
  \Crefname{figure}{Fig.}{Fig.}
  \crefname{table}{Tab.}{Tab.}
  \Crefname{table}{Tab.}{Tab.}
  \crefname{equation}{Eq.}{Eq.}
  \Crefname{equation}{Eq.}{Eq.}
  \crefname{section}{Sec.}{Sec.}
  \Crefname{section}{Sec.}{Sec.}
  \crefname{subsection}{Sec.}{Sec.}
  \Crefname{subsection}{Sec.}{Sec.}
  \crefname{subsubsection}{Sec.}{Sec.}
  \Crefname{subsubsection}{Sec.}{Sec.}
  \crefname{appendix}{Sec.}{Sec.}
  \Crefname{appendix}{Sec.}{Sec.}
  \crefname{subappendix}{Sec.}{Sec.}
  \Crefname{subappendix}{Sec.}{Sec.}
  \crefname{subsubappendix}{Sec.}{Sec.}
  \Crefname{subsubappendix}{Sec.}{Sec.}
  
}

\AtBeginEnvironment{tabular}{\rmfamily}
\AtBeginEnvironment{tabular*}{\rmfamily}

\newcommand{\experimenttable}{%
  \footnotesize
  \renewcommand{\arraystretch}{0.95}%
  \setlength{\aboverulesep}{1.5pt}%
  \setlength{\belowrulesep}{1.2pt}%
}

\makeatletter
\renewcommand{\paragraph}{%
  \@startsection{paragraph}{4}{\z@}%
    {6pt plus 1pt minus 1pt}%
    {-1em}%
    {\normalfont\normalsize\bfseries}}
\makeatother

\usepackage[normalem]{ulem}

\DeclareRobustCommand{\SuppDel}[1]{{\color{red}\sout{#1}}}
\def\confName{CVPR}
\def\confYear{2027}
\title{\name{}: Asynchronous Flow for Artist Mesh Generation, Editing, and Refinement}
\author{%
  Kaichun Qiao$^{1,2}$, Longwen Zhang$^{1,2}$, Qixuan Zhang$^{1,2}$, Taku Komura$^{3}$, Wenping Wang$^{4}$,\\
  Jingyi Yu$^{1}$, Christian Theobalt$^{5}$, Lan Xu$^{1,\dagger}$, Rui Xu$^{3,5,\dagger}$\\[0.2em]
  {\normalsize
    $^{1}$\,ShanghaiTech University\quad
    $^{2}$\,Deemos Technology Co., Ltd.\quad
    $^{3}$\,The University of Hong Kong%
  }\\
  {\normalsize
    $^{4}$\,Texas A\&M University\quad
    $^{5}$\,Max Planck Institute for Informatics%
  }\\[0.15em]
  {\small $^{\dagger}$\,Corresponding authors.}%
}

\newcommand{\KILNMAIN}{}

\begin{document}

\twocolumn[\maketitle\vspace{0em}\begin{center}
\vspace{-4.5mm}
    \includegraphics[width=\linewidth]{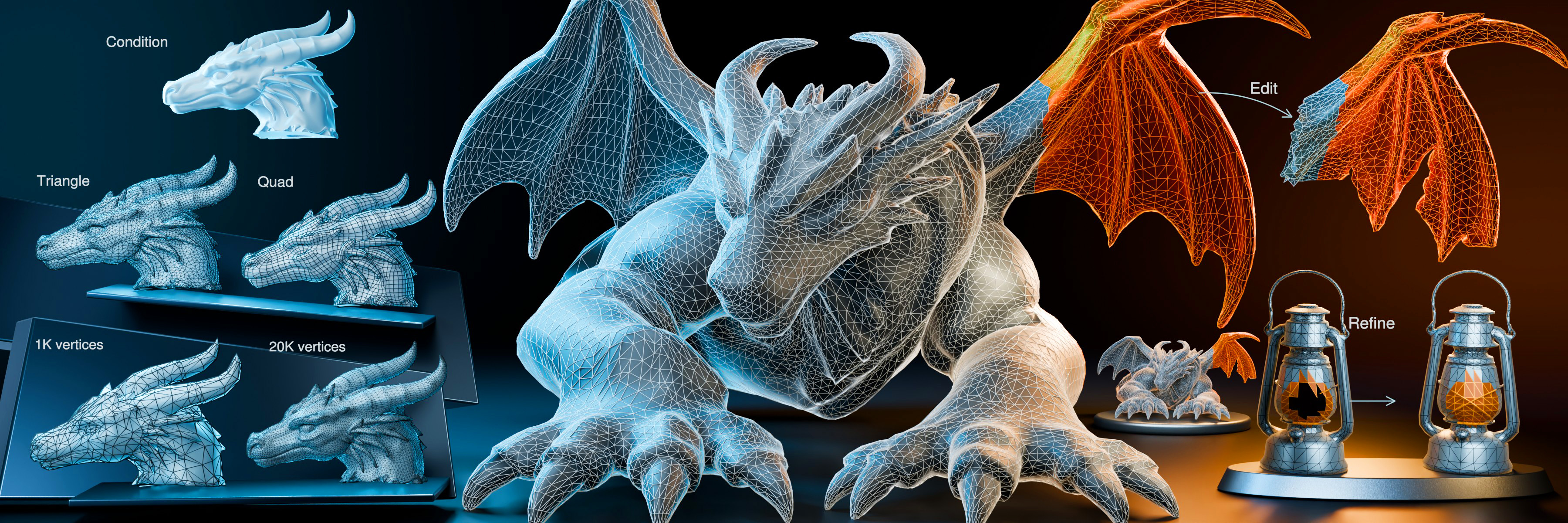}
\end{center}
\vspace{-6.5mm}
\captionof{figure} {\textbf{\name{} unifies artist-style mesh generation, local editing, and refinement.}
 Our architecture supports both triangle and quad meshes, with explicit control over vertex count, preservation of unedited regions, and local defect repair.
 All assets shown are produced by \name{}.
 }
\label{fig:teaser}
\bigbreak]
\raggedbottom

\begin{abstract}
Recent advances in artist-style mesh generation have demonstrated the potential of diffusion models built on continuous mesh representations.
Despite this progress, preserving topological integrity while enabling local editing and explicit complexity control remains challenging: existing methods frequently suffer from fragmented surfaces or holes, and offer limited support for localized revisions.
In this paper, we present \textbf{\name{}}\footnote{The name \emph{\textbf{\name{}}} evokes the final firing that turns a raw sculpture into a finished piece: our framework serves as the concluding stage of 3D generation, refining input shapes into well-tessellated, structurally sound meshes.}, a shape-conditioned framework for mesh generation, editing, and refinement.
Specifically, \textbf{\name{}} introduces a spatially addressable mesh variational autoencoder (VAE) that binds vertex latents to 3D anchors, paired with an asynchronous flow matching model that selectively updates target states while keeping the rest fixed.
This design enables seamless local editing, topology refinement, and explicit control over vertex count within a unified architecture supporting both triangle and quad meshes.
Extensive experiments demonstrate that Kiln achieves competitive geometric fidelity and superior topological quality across both
triangle and quad mesh generation tasks, while showcasing strong capabilities in local editing and topology refinement.
\end{abstract}

\section{Introduction}
\label{sec:introduction}

Artist-style meshes provide the compact geometry and structured connectivity essential for animation, games, and interactive 3D applications~\cite{meshgpt,meshanything}.
To synthesize such assets, recent diffusion-based methods have shown that continuous representations can jointly capture vertex positions and discrete connectivity, enabling parallel generation of artist-like tessellations~\cite{spacemesh,meshflow,nexus,meshyt2}.
As these generative models mature, extending them from one-shot synthesis to iterative production workflows requires the ability to revise localized regions, control mesh complexity, and preserve topological structure across edits.

\begin{figure*}[!t]
  \centering
  \includegraphics[width=\textwidth]{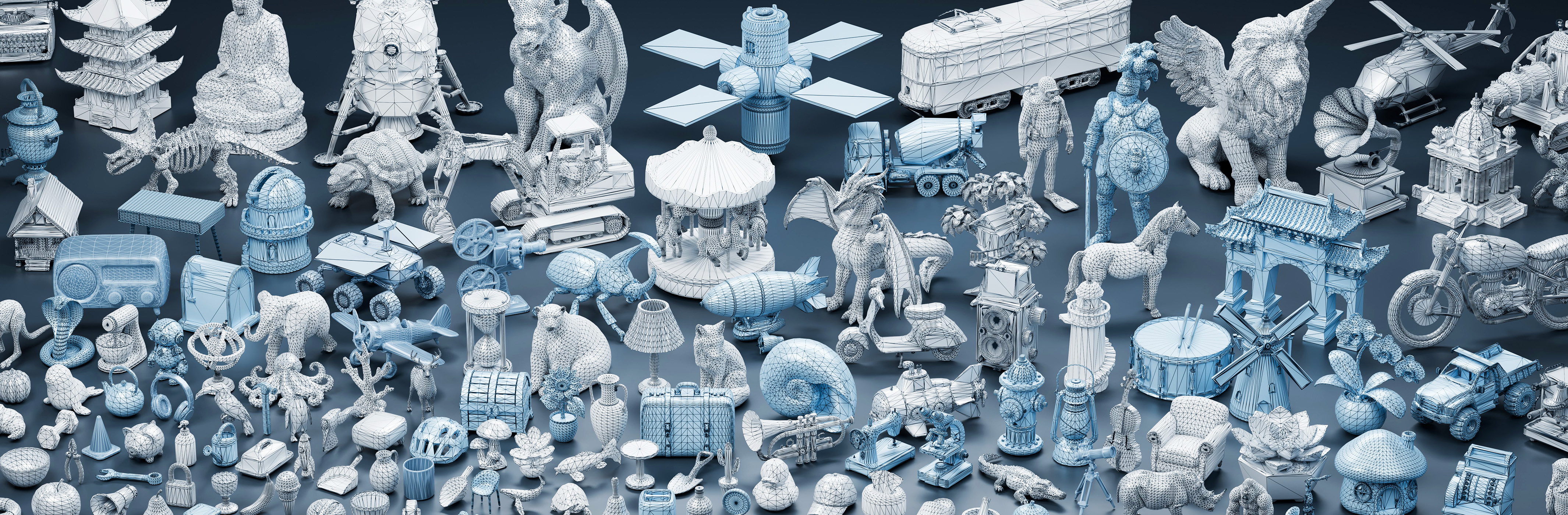}
  \vspace{-6mm}
  \caption{\textbf{Gallery of artist meshes generated by \textbf{\name{}}.} Our framework produces clean, well-structured tessellations with faithful geometry across a wide range of complex 3D shapes and object categories.}
  \vspace{-3mm}
  \label{fig:gallery}
\end{figure*}
Despite recent progress, existing generators remain difficult to integrate into these iterative workflows.
Structurally, current methods often produce fragmented surfaces, holes, or non-manifold artifacts~\cite{meshrepair,lato2}, showing that geometric accuracy alone does not ensure clean topology.
Practically, artists routinely reshape specific parts or repair local tessellations under a strict vertex budget while keeping the surrounding mesh intact.
However, most existing frameworks treat mesh synthesis as a global pass~\cite{meshgpt,meshanything,meshflow,nexus}, leaving little room for localized updates.
Even when partial regeneration is supported, adapting to modified shapes and seamlessly stitching updated regions back to the preserved boundary remain major challenges~\cite{meshfim,lato2,meshrepair}.

To overcome these limitations, our key insight is to make spatial locality native to both the latent representation and the generative process.
Guided by this principle, we present \textbf{\name{}}, a shape-conditioned framework that pairs a spatially addressable mesh variational autoencoder (VAE)~\cite{kingma2013vae} with asynchronous flow matching~\cite{flowmatching2022,combostoc}.
In our VAE, vertex latents are anchored in a coarse 3D space to jointly encode local geometry and connectivity, directly mapping spatial selections to latent states.

Building on token-wise noise scheduling~\cite{diffusionforcing,combostoc}, the flow model then selectively evolves target states while conditioning on the remaining protected states as clean, fixed context.
This formulation naturally unifies three core tasks for iterative workflows: (i)~\emph{complete-mesh generation} by evolving all latent states jointly; (ii)~\emph{local editing} by updating a selected subset under a modified shape condition; and (iii)~\emph{local refinement} by updating a selected subset under the original condition to improve local topology. 
Both local operations strictly preserve all unselected mesh regions.
Moreover, this unified architecture seamlessly supports both triangle and quad topologies.
Extensive experiments demonstrate that \textbf{\name{}} delivers competitive geometric fidelity alongside robust tessellation quality and structural integrity across diverse generation and editing benchmarks~(\cref{tab:unified-generation,tab:quad,tab:edit-compact}).
Our main contributions are summarized as follows:
\begin{itemize}
  \item We design a spatially addressable mesh VAE that binds vertex latents to coarse 3D anchors, connecting spatial regions to latent states to jointly reconstruct geometry and connectivity.
  \item We introduce an asynchronous flow matching framework that unifies complete-mesh generation, local editing, and refinement while keeping protected regions intact.
  \item We demonstrate competitive surface fidelity, accurate vertex-count control, and superior structural integrity for triangle and quad meshes, consistently producing clean, well-tessellated surfaces on evaluated benchmarks.
\end{itemize}

\section{Related Work}
\label{sec:related-work}

Converting dense 3D outputs into production-ready assets requires both structured topology and flexible local control, motivating our review of 3D synthesis, artist and quad mesh generation, shape editing, and asynchronous diffusion.

\paragraph{3D Generation.}
Early generative methods optimize per-asset 3D representations via 2D score distillation~\cite{dreamfusion,magic3d} or reconstruct surfaces from synthesized multiview images and normal maps~\cite{syncdreamer,wonder3d}, while feed-forward networks accelerate inference via direct reconstruction~\cite{lrm,triposr2024,instantmesh2024}.
Moving beyond image priors, native 3D generative models apply diffusion or flow matching over latent vector sets~\cite{shape2vecset,michelangelo,clay,triposg,hunyuan21}, deformable tetrahedral grids~\cite{meshdiffusion}, or sparse-voxel latent spaces~\cite{trellis,trellis2,direct3ds2,sparc3d,sparseflex,topomesh}.
However, converting their dense or implicit outputs into production-ready assets typically requires post-processing via geometric simplification~\cite{qem1997,pamo} or learned remeshing~\cite{nms,lightweight2025}.
We bridge this gap by directly generating and editing artist-style meshes conditioned on dense surfaces.

\paragraph{Artist Mesh Generation.}
Artist meshes compactly represent shapes with structured edge flow for editing and simulation.
Autoregressive (AR) models synthesize these discrete structures via sequential token prediction~\cite{polygen,meshgpt,meshxl,pivotmesh,llamamesh,meshllm,sato}, with subsequent methods conditioning on input surfaces~\cite{meshanything,meshanythingv2,edgerunner,rawobservation,optimusmesh} or applying reinforcement learning~\cite{deepmesh,meshrft} to improve geometric alignment and topological regularity.
Despite acceleration via sequence compression~\cite{scaling_mesh_generation,nautilus,freemesh}, localized traversals~\cite{treemeshgpt,meshweaver_cvf,meshripple,songtopology}, efficient backbones~\cite{meshtron,iflame,hifimesh}, parallelized topology decoding~\cite{fastmesh}, and speculative execution~\cite{xspecmesh,flashmesh}, sequential dependencies remain a bottleneck.

To enable parallel synthesis, diffusion and flow models operate over triangle soups~\cite{polydiff,equivariantmeshflow}, vertex-level topological embeddings~\cite{spacemesh,polyflow}, and surface fields~\cite{triflow,meshbdf}.
To improve scalability and controllability, recent frameworks compress meshes via learned autoencoders~\cite{face2026,meshcraft,meshflow,lato}, factorize vertex and topology prediction~\cite{nexus,lato2,meshcarve,taoflowforge}, or incorporate spatial scaffolds and vertex budgets~\cite{meshyt2}. In parallel, alternative paradigms decompose synthesis through part-based assembly~\cite{meshart,meshmosaic,meshrag,partdiffuser} or progressive coarse-to-fine refinement~\cite{vertexregen,armesh,mars,meshoctave}.
Within this parallel regime, \name{} introduces a spatially addressable representation that improves surface fidelity and tessellation quality while unifying full generation with local editing.

\paragraph{Quad Mesh Generation.}
Classical quadrangulation optimizes curvature-aligned directional fields and integer-grid maps~\cite{miq2009,integergrid2013,instantmeshes2015,quadriflow2018,quadwild2021}, though purely geometric objectives often miss artist-preferred edge loops.
Learning-based methods improve layout structure via neural field optimization~\cite{neurcross}, frame and singularity prediction~\cite{crossgen,neuframeq,topgen}, or chart distance fields~\cite{squadgen2026}.
Bypassing auxiliary field extraction, direct generators instead classify candidate quads~\cite{point2quad}, link vertices to face centers~\cite{quadlink}, or serialize polygons autoregressively~\cite{quadgpt,meshpro,sato,qmagic}.
Our flow-based framework similarly avoids intermediate fields to natively support quad-dominant generation with cleaner topology and tighter surface adherence.

\paragraph{Shape Editing.}
Beyond classical repair~\cite{liepa2003,attene2010} and learned subdivision~\cite{neuralsubdivision,neuralmeshrefinement,subdivar,localizedrefinement}, generative 3D editors modify shapes via multiview inpainting~\cite{instant3dit,preditor3d,cmd,omni3dedit}, masked reconstruction~\cite{maskedlrm,vecsetedit}, or latent manipulation~\cite{getmesh,voxhammer,anchorflow,editflow3d,umas3d}.
For regional editing, ShapeFusion~\cite{shapefusion} applies diffusion masking and CraftMesh~\cite{craftmesh} blends generated patches via Poisson fusion.
In latent space, NOFACE3D~\cite{noface3d} anchors boundary latents via spatial noise schedules but merges refined patches without explicit vertex-level boundary constraints, whereas SeMo3D~\cite{semo3d} unifies generation, editing, and completion via region-selective latent denoising, preserving unedited context at the token level.
Existing artist-mesh editors instead use connectivity repair~\cite{spacemesh}, two-stage vertex-then-topology updates~\cite{lato2}, sketch operations~\cite{meshpad}, or local sequence regeneration and repair~\cite{meshfim,meshrepair}.
In contrast, our continuous flow framework jointly regenerates local vertex positions and topology while explicitly preserving protected vertices and faces during editing and refinement.

\paragraph{Asynchronous Diffusion.}
Early approaches to flexible diffusion control rely on test-time resampling~\cite{repaint} or partial noising~\cite{sdedit2021} under a globally shared noise schedule.
Moving beyond uniform timesteps, sequential models introduce asynchronous schedules: Rolling Diffusion~\cite{rollingdiffusion} assigns progressively higher noise to later frames, FIFO-Diffusion~\cite{fifodiffusion} denoises frames at varied noise levels during inference with a pretrained model, and Diffusion Forcing~\cite{diffusionforcing,historyguidance} trains with independently corrupted frames to handle partially noisy histories.
Non-uniform schedules also vary noise across feature dimensions~\cite{combostoc} or spatial pixels and patches~\cite{rad,asyncdsb,asyndm,asyncpatch} so noisy regions use cleaner context.
For artist meshes, we differentiate the editable interior, boundary band, and protected context via regional time fields, recovering full-mesh generation under a shared time.

\section{Method}

\label{sec:method}

\begin{figure*}[!t]
 \centering
 \ifdefined\KILNMAIN
   \includegraphics[width=170mm]{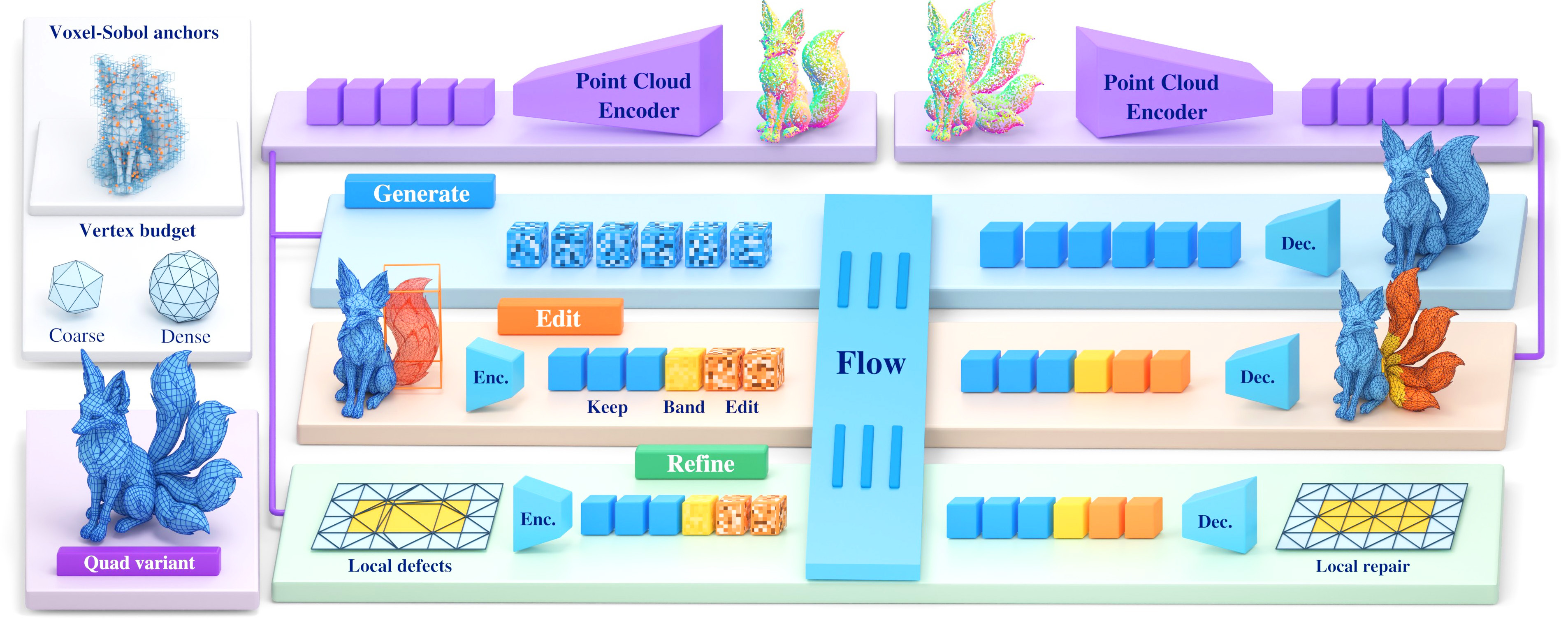}
 \else
   \fbox{\parbox[c][35mm][c]{0.9\linewidth}{\centering Method overview figure}}
 \fi
 \vspace{-3mm}
 \caption{\textbf{Overview of \name{}.}
Guided by the surface condition $c$, a flow-matching model synthesizes mesh latents at spatial anchors.
During shape editing and local refinement, it regenerates only the selected anchors while keeping protected anchors clean as context, allowing the decoder to naturally connect the updated geometry to the preserved mesh.}
\vspace{-3mm}
 \label{fig:method}
\end{figure*}

We present \name{}, a surface-conditioned framework for artist mesh generation, shape editing, and local refinement (Fig.~\ref{fig:method}).
Given a surface condition and 3D spatial anchors (Sec.~\ref{sec:data-preparation}), a mesh variational autoencoder (VAE)~\cite{kingma2013vae} maps meshes to continuous anchor states and decodes vertex occupancy, positions, and connectivity (Sec.~\ref{sec:representation}).
A conditional flow-matching model then evolves these states under an asynchronous time field, synthesizing all states for complete generation or updating spatially selected states while keeping the rest clean to connect naturally to the preserved boundary (Sec.~\ref{sec:async-flow}).

\subsection{Surface Conditions and Spatial Anchors}

\label{sec:data-preparation}

\begin{figure}[!t]
 \centering
 \includegraphics[width=\linewidth]{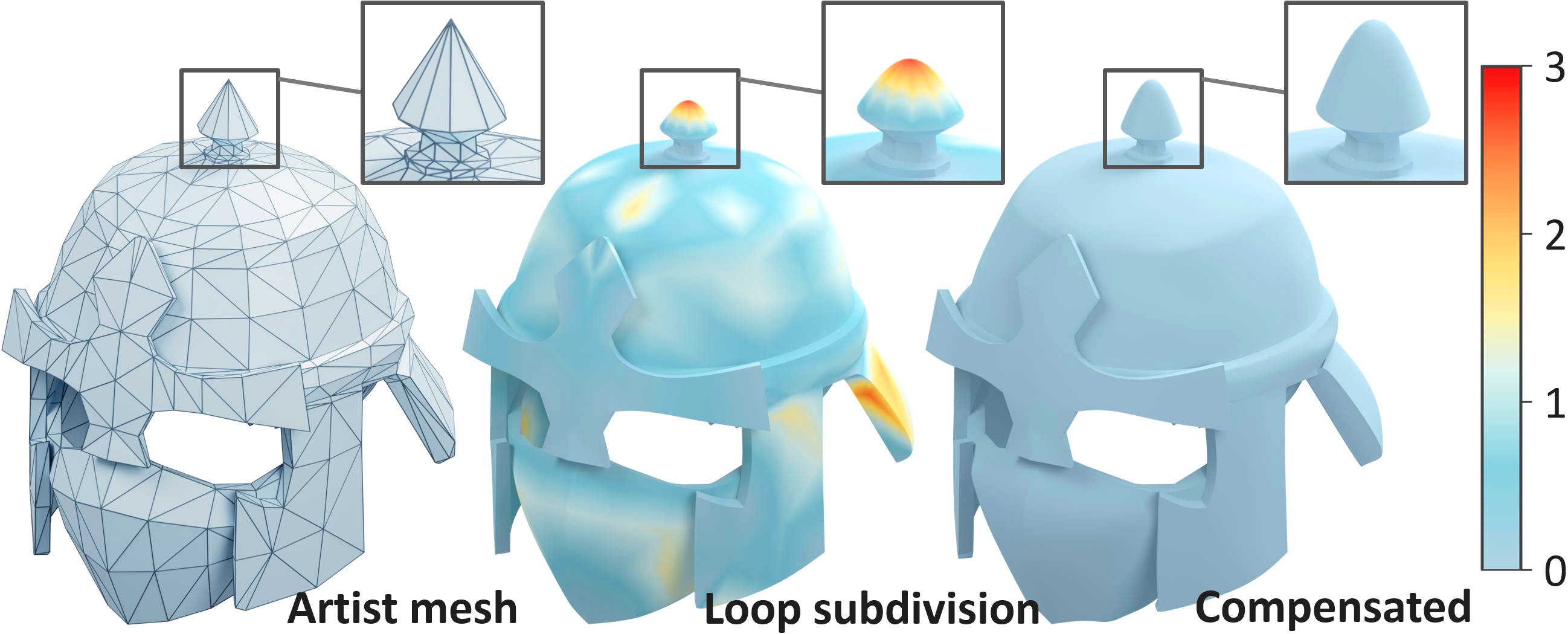}
 
 \vspace{-2mm}
 \caption{\textbf{Shrinkage compensation.}
An artist mesh and its Loop surfaces before and after compensation, colored by the displacement between the original vertices and their limit positions.}
 \label{fig:loop-shrinkage}
\end{figure}

\paragraph{Surface conditions.}

Training targets are artist meshes $M=(V,F)$ with sparse, structured tessellations, whereas the meshes provided as inference inputs, 
typically generated by image-to-3D models~\cite{triposr2024,instantmesh2024,clay,trellis,hunyuan21} or surface reconstruction methods~\cite{neus,volsdf,neuralangelo,sugar,gaussian2d}, exhibit dense, near-uniform triangulations~\cite{marchingcubes1987}.
Sampling points directly from $M$ would expose flat facets and sharp normal discontinuities, leaking its edge layout and creating an inference domain gap.
To suppress these topological cues, we construct a smooth proxy surface $H$ via Loop subdivision~\cite{loop1987smooth} with shrinkage compensation~\cite{deng2012loopinterpolation}. Specifically, we treat original vertices as limit-surface targets and invert the Loop limit weights to solve for corrected control points before subdivision, counteracting contraction and preserving the original vertex positions in $H$ (Fig.~\ref{fig:loop-shrinkage}).
After normalizing both training and inference meshes by centering their bounding boxes at the origin and uniformly scaling them into $[-1,1]^3$, we sample $81,920$ points and normals from $H$ and encode them with the Hunyuan3D-2.1~\cite{hunyuan21} shape encoder to obtain the surface condition $c$ .

\paragraph{Spatial anchors.}

A key challenge in representing artist meshes is their irregular vertex counts and spatial distributions, whereas local editing and refinement demand a spatially addressable representation.
To decouple spatial addressing from fine geometric detail, we represent candidate vertex locations using $L$ coarse 3D anchors $A=\{a_i\}_{i=1}^{L}$ as spatially organized slots, while delegating detailed shape cues to the surface condition $c$ (see \cref{sec:anchor-condition-diagnostic} for an ablation against surface-sampled anchors). 
Specifically, we rasterize the proxy surface $H$ into a $64^3$ occupancy grid, distribute $L$ anchors evenly across occupied voxels, and draw their exact coordinates within each voxel via a low-discrepancy Sobol sequence~\cite{sobol1967,matousek1998}.
We then formulate vertex-to-anchor pairing as a distance-minimizing assignment problem~\cite{kuhn1955assignment}.
To bypass the $\mathcal{O}(L^3)$ bottleneck of exact solvers at large $L$, we apply fast batched iterative matching over local candidates (\cref{sec:supp-anchor-matching}) to assign each vertex to a distinct anchor, leaving unmatched anchors vacant.

\subsection{Spatially Addressable Mesh VAE}

\label{sec:representation}

\paragraph{Encoding.}

\label{sec:mesh-encoding}

\begin{figure}[!t]
 \centering
 \includegraphics[width=\linewidth]{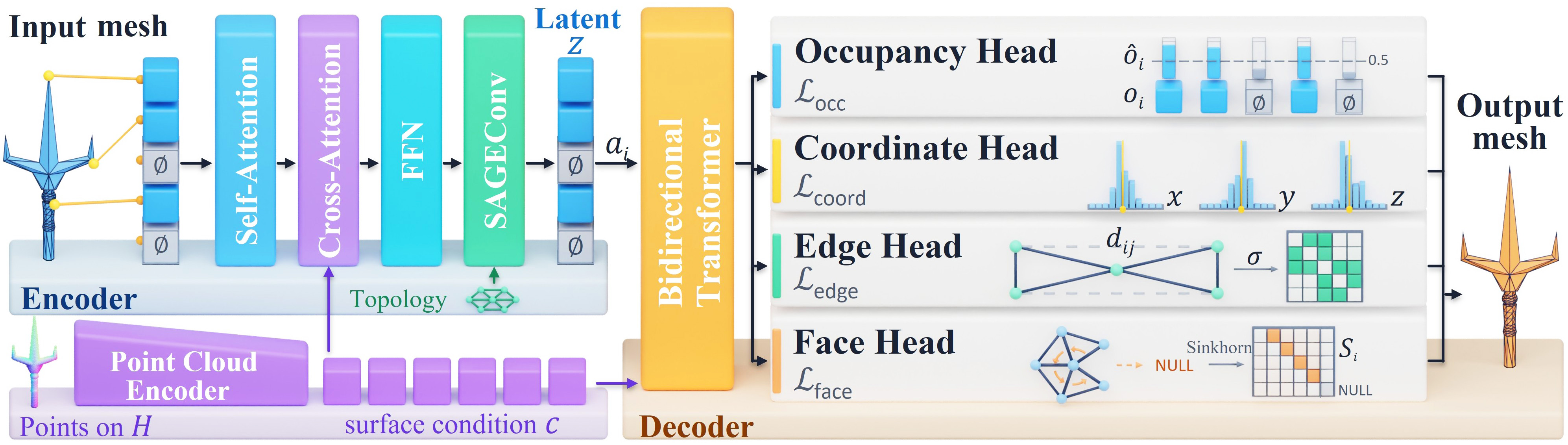}
 
\vspace{-3mm}
 \caption{\textbf{Mesh VAE architecture.}
The encoder integrates vertex attributes with mesh connectivity and the surface condition $c$, while the decoder reconstructs anchor occupancy, vertex coordinates, edges, and face successors.}
 \label{fig:mesh-vae}
\end{figure}

Under our spatial anchor assignment, we encode each mesh vertex into a 24-dimensional latent $z_i$ at its matched anchor, while unmatched anchors share a learned empty latent.
Following Nexus~\cite{nexus}, vertex features initialized from coordinates and normals are refined across encoder layers combining self-attention, cross-attention to surface condition $c$~\cite{vaswani2017attention}, and SAGEConv~\cite{graphsage2017} along mesh edges (Fig.~\ref{fig:mesh-vae}).

\paragraph{Occupancy and geometry decoding.}

\label{sec:mesh-decoding}

Given latent $z_i$, anchor $a_i$, and condition $c$, the decoder first predicts an occupancy probability $\hat o_i$ supervised against binary labels $o_i$ with binary cross-entropy ($\mathcal L_{\mathrm{occ}}$; \cref{sec:supp-vae-heads}).
At inference, vacant anchors are discarded to establish active vertices, whereas training supervises geometry and connectivity solely on occupied anchors.
For each occupied anchor, the decoder predicts discrete distributions over 512 bins along each of the $x$, $y$, and $z$ axes.
We supervise these distributions with soft-label cross-entropy against a target truncated Gaussian centered at the ground-truth coordinate:
\begin{equation}
 \mathcal L_{\mathrm{coord}}
 =-\mathbb E\!\left[\sum_{k=1}^{512}q_k\log p_k\right],
 \label{eq:expected-ce}
\end{equation}
where $p_k$ and $q_k$ denote the predicted and target probabilities for bin $k$.
Consistent with this soft target, we recover continuous coordinates via the expected bin center $\sum_k p_k b_k$ rather than a hard argmax, preserving fine-grained, sub-bin precision (see Sec.~\ref{sec:coordinate-decoding} for ablation).

\paragraph{Edge and face decoding.}

Following Nexus~\cite{nexus}, the decoder projects each active-vertex feature into spatial and temporal embeddings $\mathbf s_i,\mathbf t_i\in\mathbb R^{32}$ to model pairwise connectivity via the spacetime interval~\cite{spacemesh}:
\begin{equation}
 d_{ij}=\|\mathbf s_i-\mathbf s_j\|_2^2-\|\mathbf t_i-\mathbf t_j\|_2^2.
 \label{eq:spacetime-edge}
\end{equation}
Applying the sigmoid $\sigma$ yields edge probabilities $\hat e_{ij}=\sigma(d_{ij})$ for supervision, while thresholding at $d_{ij}>0$ extracts the predicted edge graph at inference to constrain subsequent half-edge routing and face assembly.
Because edges are sparse, uniform binary cross-entropy (BCE) would let abundant non-edges dominate training.
We therefore partition vertex pairs into true positives ($\mathrm{TP}$), true negatives ($\mathrm{TN}$), false positives ($\mathrm{FP}$), and false negatives ($\mathrm{FN}$) against ground-truth labels $e_{ij}$, equally weighting each group's mean BCE so that $\mathrm{TN}$ pairs do not overwhelm penalties on spurious or missed edges (\cref{sec:supp-vae-heads}):
\begin{equation}
 \mathcal L_{\mathrm{edge}}
 =\frac14\sum_{g\in\{\mathrm{TP,TN,FP,FN}\}}
 \mathbb E_{(i,j)\in g}
 \!\left[\operatorname{BCE}(\hat e_{ij},e_{ij})\right].
 \label{eq:edge-loss}
\end{equation}

\begingroup
\setlength{\intextsep}{3pt}
\setlength{\columnsep}{7pt}

\begin{wrapfigure}{r}{22mm}
 \centering
 \includegraphics[width=\linewidth]{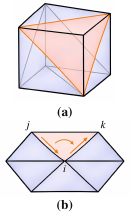}
\end{wrapfigure}

However, edge connectivity alone does not uniquely determine mesh faces, as in inset figure (a), where three surface edges form a cycle whose interior is not a face (red).
To resolve this ambiguity, we predict directed half-edge successors~\cite{spacemesh,meshyt2}, where a corner $j\!\to\!i\!\to\!k$ continues $j\!\to\!i$ along $i\!\to\!k$, shown in orange in (b).
At each vertex $i$, the decoder applies Sinkhorn normalization~\cite{sinkhornknopp1967} to predicted neighbor scores, forming a transition matrix $S_i$ of continuation probabilities $S_i(j,k)$ with a $\mathrm{NULL}$ state for open boundaries.
During training, the ground-truth faces around vertex $i$ induce a successor mapping $\pi_i$ that pairs each incoming neighbor $j$ with its outgoing neighbor, while reversing the local loop traversal direction yields the inverse mapping $\pi_i^{-1}$.
To remain invariant to orientation, we evaluate the likelihood of each traversal direction under the predicted transition probabilities $S_i(j,k)$:
\begin{equation}
 P_i^+=\prod_j S_i(j,\pi_i(j)),\qquad
 P_i^-=\prod_j S_i(j,\pi_i^{-1}(j)).
 \label{eq:face-likelihood}
\end{equation}
Marginalizing over both orientations and normalizing by the total number of supervised assignments $D$ yields the face reconstruction loss (detailed in \cref{sec:supp-vae-heads}):
\begin{equation}
 \mathcal L_{\mathrm{face}}
 =-\frac1D\sum_i
 \log\frac{P_i^++P_i^-}{2}.
 \label{eq:face-loss}
\end{equation}
Because half-edge successors define face corners regardless of polygon degree, we can train a separate model on native quadrilateral meshes with the same representation.
At inference, decoding prioritizes four-edge cycles and uses triangles only for residual three-edge cycles.

\endgroup

\paragraph{Training objective.}

\label{sec:vae-objective}

\label{sec:quad}

We regularize each occupied anchor's Gaussian posterior toward a standard normal prior via $\mathcal L_{\mathrm{KL}}$, combining this term with the reconstruction losses:
\begin{equation}
 \begin{aligned}
 \mathcal L_{\mathrm{VAE}}
 ={}&\lambda_{\mathrm{occ}}\mathcal L_{\mathrm{occ}}
 +\lambda_{\mathrm{coord}}\mathcal L_{\mathrm{coord}}\\
 &+\lambda_{\mathrm{edge}}\mathcal L_{\mathrm{edge}}
 +\lambda_{\mathrm{face}}\mathcal L_{\mathrm{face}}
 +\lambda_{\mathrm{KL}}\mathcal L_{\mathrm{KL}}.
 \end{aligned}
 \label{eq:vae-objective}
\end{equation}
Loss weights and VAE training settings are detailed in \cref{sec:supp-vae-training}.

\subsection{Asynchronous Flow over Anchor States}

\label{sec:async-flow}

\paragraph{Anchor states and vertex budget.}

\label{sec:whole-mesh-generation}

With the VAE frozen, we standardize the 24-dimensional anchor latents $z_i$, comprising encoded latents for matched vertices and the shared empty latent for unmatched anchors, to form the flow-matching target $Y^*\in\mathbb R^{L\times24}$.
This design relies on the decoder to predict occupancy directly without an extra flow-state dimension.
A vertex budget $B$, given by the ground-truth vertex count during training and the requested budget at inference, conditions the flow model and determines the sequence capacity $L$.
Specifically, we round $1.25B$ upward to the smallest allowable preset capacity (7 levels, capped at $20\text{K}$; \cref{sec:supp-anchor-capacity}), reducing computation on small budgets while accommodating spatially uneven vertex densities.
Because distinct budgets can share the same capacity $L$, the realized vertex count remains soft and emerges jointly from $c$, the budget conditioning $B$, and the decoded occupancy probabilities $\hat o_i$.

\begin{figure*}[!t]
\captionsetup{skip=3pt}
 \centering
 \includegraphics[width=\textwidth]{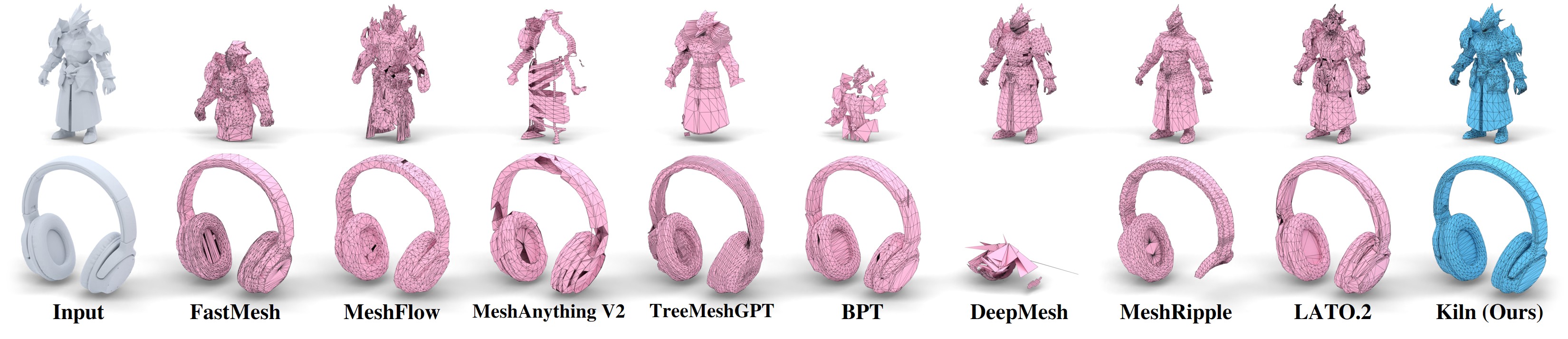}
 \vspace{-6mm}
 \caption{\textbf{Triangle-mesh generation.}
Compared with existing state-of-the-art baselines, \name{} reconstructs more intact geometries with cleaner, better-organized mesh connectivity.}
\vspace{-4mm}
 \label{fig:triangle-comparison}
\end{figure*}

\paragraph{Asynchronous Flow matching.}

\label{sec:async-training}

To generate these anchor states, our flow-matching model predicts a velocity field transporting Gaussian noise $\epsilon$ to the target states $Y^*$.
However, a shared scalar time $t$ subjects all anchors to the same noise level, preventing the model from keeping a reference mesh fixed while updating a local region.
Following token-wise noise scheduling~\cite{diffusionforcing,combostoc}, we introduce an anchor-wise time field $T\in[0,1]^L$ along the linear path~\cite{flowmatching2022}:
\begin{equation}
 Y_{T,i}=(1-T_i)\epsilon_i+T_iY_i^*,\qquad
 u_i=Y_i^*-\epsilon_i,
 \label{eq:async-path}
\end{equation}
where $\epsilon_i\sim\mathcal N(0,I)$ and $u_i$ denotes the target velocity.
Setting $T_i=t$ across all anchors recovers complete-mesh generation, whereas setting $T_i=1$ keeps an anchor clean as fixed context.

We then parameterize the velocity flow matching network $v_\theta(Y_T,T;c,A,B)$ as a Transformer over anchor tokens conditioned on each state $Y_{T,i}$, anchor position $a_i$, time $T_i$, and vertex budget $B$, exchanging context via self-attention and cross-attention to the surface condition $c$.
With the mesh VAE and surface encoder frozen, we train the velocity network in two stages: synchronous complete-mesh pretraining followed by mixed-stage adaptation that jointly samples complete-mesh synchronous and regional asynchronous time fields (detailed training configurations in \cref{sec:supp-time-field-details}).
For each training sample, we draw a base time $s$ and set $T_i=s$ across all anchors for complete-mesh steps.
For regional steps, we sample a local box as the editable region $U$, leave outside anchors as the protected set $K$, and designate the $15\%$ of anchors in $U$ closest to $K$ as an internal boundary band $R\subset U$.
We assign $T_i=s$ to the editable interior $U\setminus R$, $T_i=(1+s)/2$ to the boundary band $R$, and $T_i=1$ to the protected anchors in $K$, providing less corrupted transitional states near the boundary so that regenerated interior tokens can smoothly adapt to the fixed surrounding context.
In both settings, we supervise the predicted velocity only on noisy anchors $\Omega=\{i:T_i<1\}$:
\begin{equation}
 \mathcal L_{\mathrm{async}}
 =\mathbb E_{Y^*,\,\epsilon,\,T}\!\left[\frac1{|\Omega|}\sum_{i\in\Omega}
 \left\|v_{\theta,i}(Y_T,T;c,A,B)-u_i\right\|_2^2\right].
 \label{eq:async-loss}
\end{equation}

\paragraph{Shape editing and local refinement.}

\label{sec:local-support}

At inference, given a reference state $Y^{\mathrm{ref}}$, we expand the user selection into an editable region $U$, defining the protected set $K$ and internal boundary band $R\subset U$ as above.
Keeping protected states fixed at $Y_K^{\mathrm{ref}}$, we update $U$ under the time field
\begin{equation}
 T_i=t\ (i\in U\setminus R),\; T_i=\tfrac{1+t}{2}\ (i\in R),\; T_i=1\ (i\in K).
 \label{eq:regional-flow}
\end{equation}
For shape editing, we extract the new shape condition $c^*$ from the modified surface, retain reference anchors in $R$ and $K$, and resample interior anchors in $U\setminus R$ under vertex budget $B$ (Sec.~\ref{sec:data-preparation}), initializing their latents from pure Gaussian noise at $t_0=0$.
For local refinement, we instead reuse the original condition $c$ and existing anchors, partially noising interior latents to a starting time $t_0>0$ following SDEdit~\cite{sdedit2021}:
\begin{equation}
 Y_{t_0,U\setminus R}=(1-t_0)\epsilon_{U\setminus R}+t_0Y_{U\setminus R}^{\mathrm{ref}}.
 \label{eq:refinement-start}
\end{equation}

In both operations, we initialize the boundary band $R$ from its reference latents via the same noise interpolation at time $(1+t_0)/2$, which corresponds to an equal mixture of reference state and Gaussian noise when $t_0=0$.
When integrating from $t_0$ to $1$, we scale predicted velocities on $R$ by $1/2$ to match the time derivative of its schedule while clamping $K$ to $Y_K^{\mathrm{ref}}$ (\cref{sec:supp-local-operation-details}).
Decoding the full anchor layout then predicts the updated region alongside its boundary connections to $K$.
To preserve the protected geometry exactly, we restore the original reference vertices and faces strictly inside $K$ while adopting all decoded vertices in $U$ and any decoded faces incident to $U$, including faces bridging $R$ and $K$.
By re-evolving partially noised latents under the clean surface condition $c$ and surrounding context $K$, local refinement jointly regularizes vertex positions and rewires face connectivity within $U$, closing open boundary holes, untangling disordered edge flows while seamlessly bridging back to the preserved mesh (Sec.~\ref{sec:edit-evaluation}, \cref{sec:supp-local-operation-details}).

\begingroup
\raggedbottom
\setlength{\floatsep}{6pt plus 1pt minus 1pt}
\setlength{\textfloatsep}{9pt plus 2pt minus 2pt}
\setlength{\intextsep}{7pt plus 1pt minus 1pt}
\setlength{\dblfloatsep}{6pt plus 1pt minus 1pt}
\setlength{\dbltextfloatsep}{9pt plus 2pt minus 2pt}
\captionsetup{skip=3pt}
\section{Experiments}
\label{sec:experiments}

We evaluate \name{} as a unified framework across triangle mesh generation, vertex-count control, local editing and refinement, and native quad generation, complemented by image-conditioned comparisons with commercial methods.
A single set of weights supports all triangle-mesh tasks, whereas native quad generation uses separate weights.
Controlled ablations in the supplementary material further support our key design choices, including regional training, the asynchronous boundary band, and cross-entropy loss for coordinates, alongside surface-conditioning analyses and full implementation and evaluation protocols 
(\cref{sec:supp_arch,sec:supp_eval,sec:supp-ablation}).
We evaluate performance across surface fidelity, mesh topology, vertex-count accuracy, boundary continuity, and runtime; exact metric definitions and normalization conventions are detailed in \cref{sec:supp-metrics}.
\label{sec:experimental-setup}
\label{sec:implementation-details}

\subsection{\texorpdfstring{Mesh Generation}{Mesh Generation}}
\label{sec:generation-results}

\paragraph{Comparison setup.}
We benchmark complete-mesh generation against eight surface-conditioned artist-mesh generators: FastMesh~\cite{fastmesh}, BPT~\cite{scaling_mesh_generation}, TreeMeshGPT~\cite{treemeshgpt}, DeepMesh~\cite{deepmesh}, MeshAnything V2~\cite{meshanythingv2}, MeshFlow~\cite{meshflow}, MeshRipple~\cite{meshripple}, and LATO.2~\cite{lato2}. The benchmark comprises 250 test shapes spanning artist meshes and TRELLIS.2~\cite{trellis2} generated meshes (\cref{fig:supp-generation-inputs}), with each baseline retaining its native output complexity budget.
Notably, we evaluate both official variants of MeshRipple~\cite{meshripple} configured with 10k and 20k face budgets. The 20k variant incorporates Native Sparse Attention to scale generation to higher face counts.
To evaluate whether generated outputs are both geometrically accurate and structurally sound, we jointly examine surface geometric fidelity, connectivity quality, and inference efficiency in \cref{tab:unified-generation}, 
where CD and HD95 are reported in $10^{-2}$ units, $\text{NM\%}$ denotes the percentage of non-manifold edges, and valence irregularity (Val. irr.) measures the fraction of vertices deviating significantly from ideal six-valent connectivity.
In particular, \emph{SL/1k} measures the count density of small open boundary loops rather than beneficial edge loops; complete metric definitions and benchmark protocols are detailed in \cref{sec:supp-metrics} and \cref{sec:supp_eval}.

\begin{table}[!t]
\captionsetup{skip=3pt}
\caption{\textbf{Surface-conditioned triangle mesh generation.}
Compared with SOTA methods on 250 test inputs.
}

\label{tab:unified-generation}
 \centering
 \experimenttable
 \relax%
 \setlength{\tabcolsep}{2.2pt}
 \renewcommand{\arraystretch}{0.95}
 \resizebox{\columnwidth}{!}{%
 \begin{tabular}{l|ccc|ccc|c}
  \toprule
 & \multicolumn{3}{c|}{\textbf{Surface fidelity}} & \multicolumn{3}{c|}{\textbf{Mesh structure}} & \textbf{Efficiency}\\
\midrule
\textbf{Method} & \textbf{CD$\downarrow$} & \textbf{HD95$\downarrow$} & \textbf{F1$\uparrow$} & \textbf{Val. irr.$\downarrow$} & \textbf{NM\%$\downarrow$} & \textbf{SL/1k$\downarrow$} & \textbf{Time (s)$\downarrow$}\\
  \midrule
  FastMesh & 1.895 & 4.490 & 0.482 & 0.804 & 79.797 & 1.410 & 37.57\\
  BPT & 1.716 & 3.543 & \textbf{0.556} & 0.209 & 1.352 & 1.731 & 212.96\\
  TreeMeshGPT & 2.022 & 5.214 & 0.496 & \textbf{0.197} & \textbf{0.739} & 4.516 & 145.45\\
  DeepMesh & 15.986 & 37.179 & 0.135 & 0.224 & 1.684 & 1.247 & 695.91\\
  MeshAnything V2 & 2.377 & 6.126 & 0.396 & 0.332 & 4.134 & 2.978 & 123.44\\
  MeshFlow & 1.385 & 2.831 & 0.389 & 0.279 & 13.569 & 0.890 & \underline{\textbf{4.61}}\\
  MeshRipple (10k) & 2.834 & 6.951 & 0.456 & 0.218 & 1.663 & 3.653 & 846.90\\
  MeshRipple (20k) & 2.464 & 5.856 & 0.458 & 0.270 & 1.835 & 2.278 & 1417.69\\
  LATO.2 & \textbf{1.076} & \textbf{2.079} & 0.539 & 0.369 & 30.819 & \textbf{0.210} & \textbf{5.38}\\
  \name{} (Ours) & \underline{\textbf{0.782}} & \underline{\textbf{1.470}} & \underline{\textbf{0.743}} & \underline{\textbf{0.113}} & \underline{\textbf{0.000}} & \underline{\textbf{0.152}} & 14.81\\
  \bottomrule
 \end{tabular}
 }
  \vspace{-2mm}
\end{table}

\paragraph{Results.}
As summarized in \cref{tab:unified-generation}, although some lighter continuous baselines achieve faster inference, \name{} delivers stronger shape fidelity and topological quality across all the evaluated methods.
These quantitative findings align closely with the visual comparisons in \cref{fig:triangle-comparison,fig:supp_tri_gallery}, where \name{} recovers complete shapes, fine part structure, and coherent connectivity while several baselines exhibit missing parts, disconnections, or shape degradation.
It is worth noting that, owing to our well-structured network and spatially grounded representation, \name{} exhibits no non-manifold defects across the entire test suite, 
without any post-processing.

\begin{table}[!t]
\caption{\textbf{Vertex-count control on 250 test inputs.}
\name{} achieves more accurate vertex count control while maintaining better geometric consistency across resolutions.
}

\label{tab:vertex-control}
 \centering\experimenttable
 \setlength{\tabcolsep}{1.5pt}
 \begin{tabular}{@{}l|ccccc@{}}
 \toprule
\textbf{Method} & \textbf{V-err.\%$\downarrow$} & \textbf{P95\%$\downarrow$} & \textbf{$S_{5\%}$\%$\uparrow$} & \textbf{CD$\downarrow$} & \textbf{NM\%$\downarrow$}\\
 \midrule
 MeshFlow (resized) & 48.690 & 61.500 & 0.000 & 2.902 & 54.350\\
 MeshFlow (fixed) & \textbf{5.620} & \textbf{10.480} & \textbf{54.200} & 1.326 & \textbf{12.780}\\
 LATO.2 & 12.803 & 28.218 & 33.280 & \textbf{1.055} & 30.605\\
 \name{} (Ours) & \underline{\textbf{0.650}} & \underline{\textbf{1.340}} & \underline{\textbf{100.000}} & \underline{\textbf{0.977}} & \underline{\textbf{0.000}}\\
 \bottomrule
 \end{tabular}
\end{table}

\begin{figure}[!t]
 \centering
 \includegraphics[width=\columnwidth]{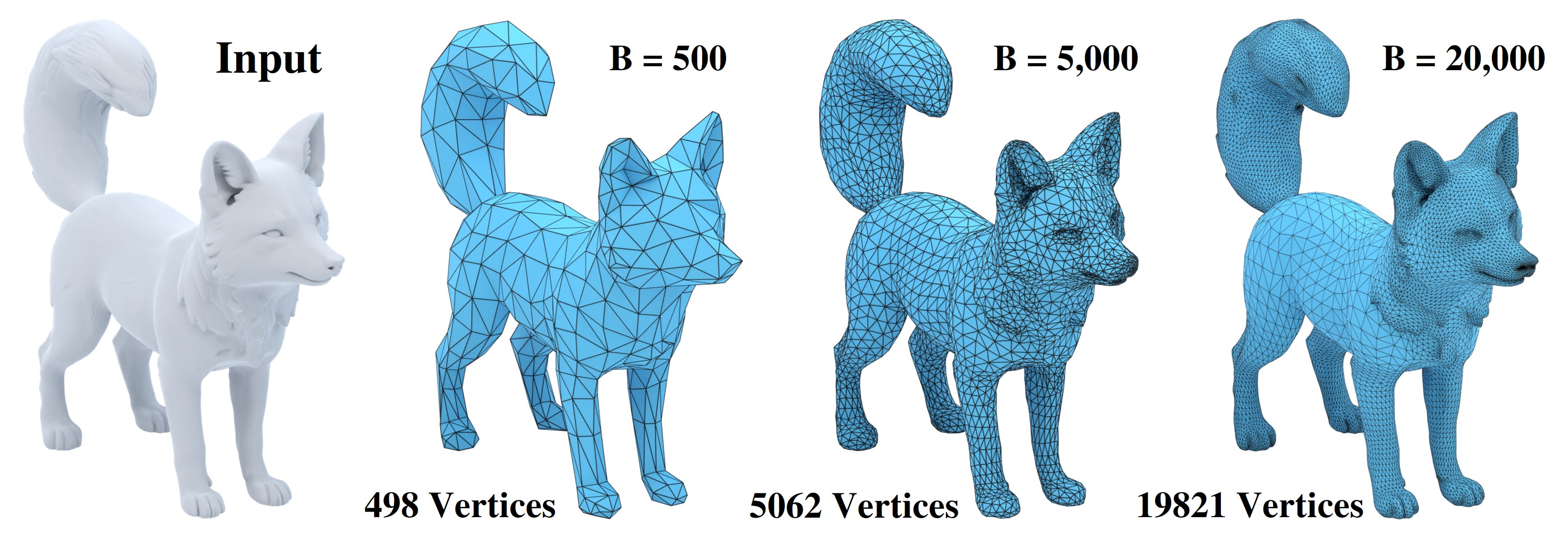}
 \vspace{-6mm}
 \caption{\textbf{Vertex-count control from a shared input surface.}
The input surface is shown on the left, followed by meshes generated at the requested vertex budgets $B$ alongside their actual vertex counts.
}
 \label{fig:vertex-count-qualitative}
\end{figure}

\paragraph{Vertex-count control.}
We compare \name{} with MeshFlow~\cite{meshflow} and LATO.2~\cite{lato2}, the two open-source baselines in our evaluation that support explicit vertex-count conditioning. For MeshFlow, we evaluate both fixed-length and proportionally resized configurations (see \cref{sec:supp-count-control} for details).
In \cref{tab:vertex-control}, \name{} adheres more accurately to target budgets while better preserving surface geometry and manifold structure.
\Cref{fig:vertex-count-qualitative} further demonstrates the controllable output complexity of our method, producing progressively denser, well-structured meshes from the same input.

\begin{figure}[!t]
 \centering
 \includegraphics[width=0.9\columnwidth]{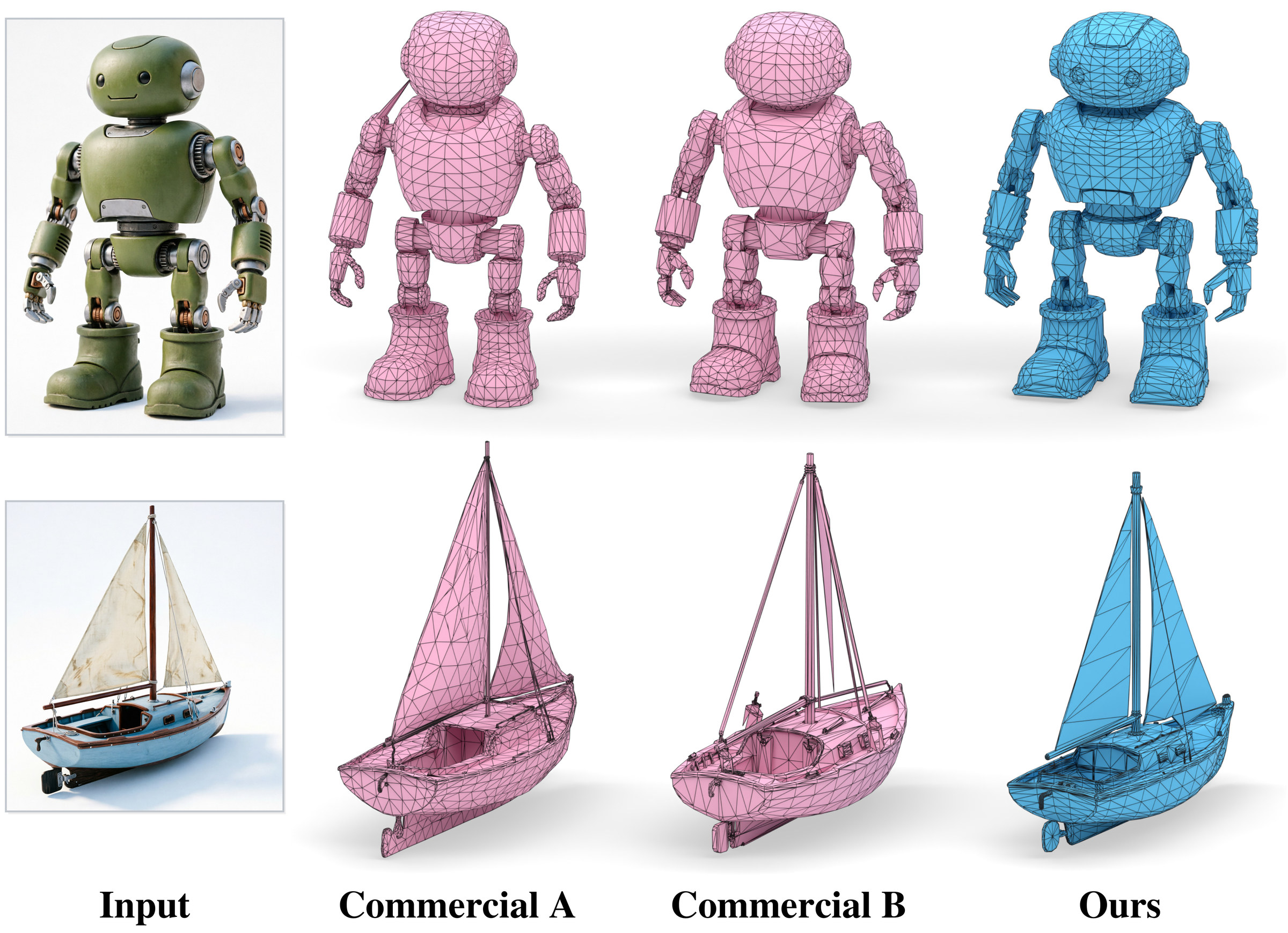}
 \caption{\textbf{Image-conditioned commercial comparisons.} All methods are queried with a shared budget of 3,000 vertices.}
 \label{fig:image-conditioned-comparison}
\end{figure}

\begin{figure}[!t]
 \centering
 \includegraphics[width=0.85\columnwidth]{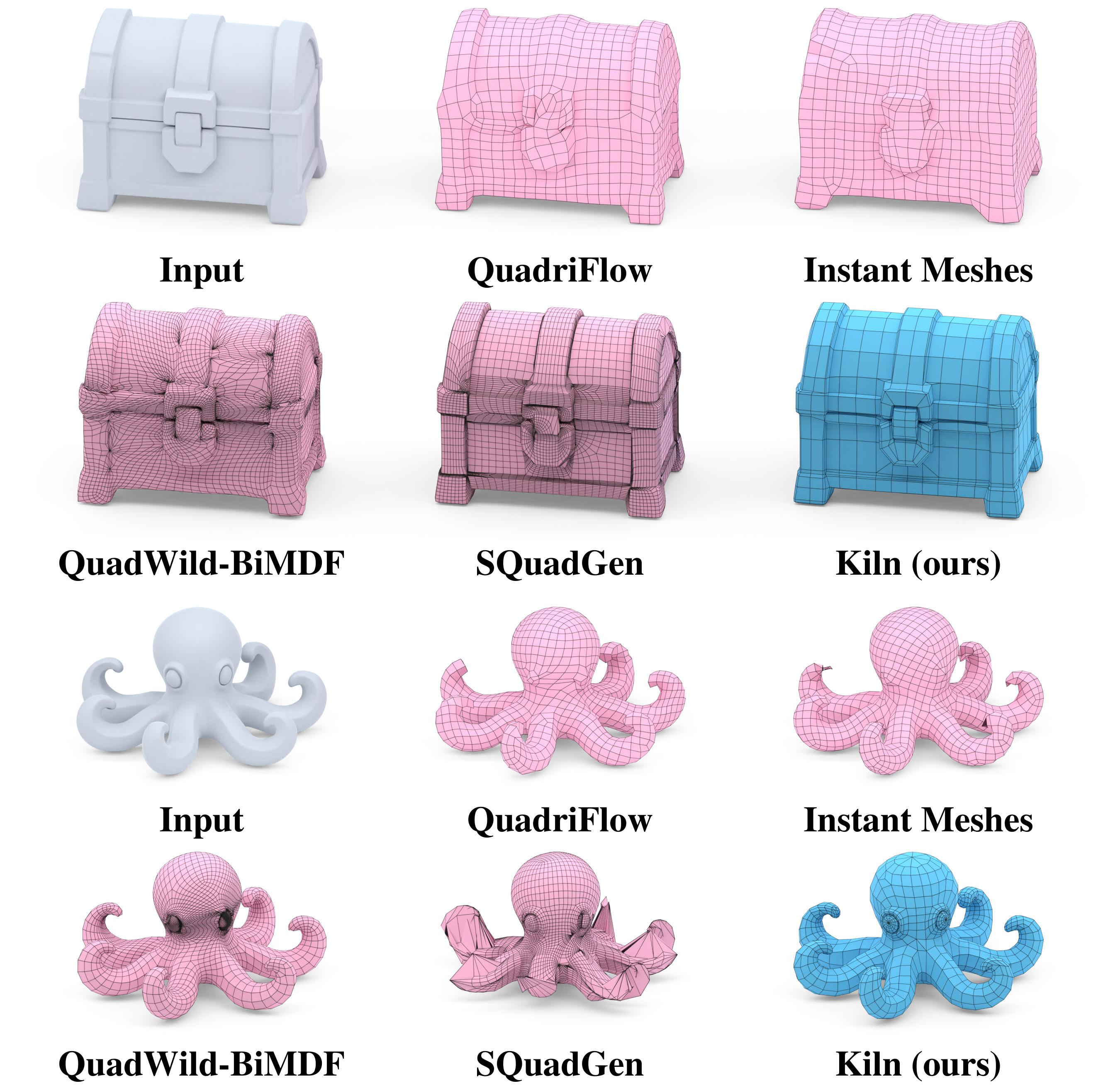}
 \vspace{-1mm}
 \caption{\textbf{Quad-mesh comparison.}
Compared with existing quad remeshing and generation methods, \name{} better preserves thin structures and produces more regular, well-aligned quad layouts.
}
 \label{fig:quad-comparison}
\end{figure}

\paragraph{Image-conditioned generation.}
We compare image-conditioned generation with two commercial models, denoted Commercial A and Commercial B, both of which accept only image inputs. 
Because Kiln is surface-conditioned, we first use TRELLIS.2~\cite{trellis2} to convert each input image into an intermediate surface before synthesizing the final mesh (\cref{sec:supp-image-comparison}).
As shown in \cref{fig:image-conditioned-comparison}, although Commercial A and Commercial B operate directly on the input images, \name{} still achieves better-preserved geometric details, sharper thin features, and cleaner, more coherent tessellations across the reconstructed surfaces.

\paragraph{Native Quad Generation.}
\label{sec:quad-results}
For native quad mesh generation, we compare \name{} against the classical quad remeshers Instant Meshes~\cite{instantmeshes2015}, QuadriFlow~\cite{quadriflow2018}, and QuadWild-BiMDF~\cite{quadwild2021}, and the generative baseline SQuadGen~\cite{squadgen2026}.
As shown in \cref{tab:quad}, \name{} achieves higher surface fidelity and superior corner-angle regularity than all competing methods, where $\text{AngOut\%}$ denotes the percentage of quad corner angles outside $[30^\circ, 150^\circ]$.
Qualitatively, \cref{fig:quad-comparison,fig:supp_quad_gallery} demonstrate that our approach better preserves slender structures and distinct part geometry while producing clean, well-aligned quad layouts.
Additional quad-dominant generation results across diverse shapes and detailed evaluation settings are provided in \cref{fig:supp-quad-showcase} and \cref{sec:supp_eval}, respectively.

\subsection{Editing and Refinement}
\label{sec:edit-evaluation}

Beyond complete mesh generation, we evaluate \name{} on local shape editing toward a modified target surface and local refinement of mesh connectivity under the original surface.
Both settings update an expanded editable region while keeping the protected region $K$ fixed and maintaining valid connectivity across the interface.
We assess regional fidelity and interface completeness (IC\%, the fraction of expected seam edges shared by exactly one preserved face and one regenerated face), detailing benchmark protocols and metric criteria in \cref{sec:supp-metrics,sec:supp-local-editing}.

\begin{table}[!t]
\vspace{-1mm}
\caption{\textbf{Native quad generation on 250 test inputs.}
\name{} demonstrates comprehensive superiority over these state-of-the-art quad mesh generation methods.}

\label{tab:quad}

 \centering\experimenttable
 \setlength{\tabcolsep}{2pt}
 \begin{tabular}{l|ccc}
\toprule
\textbf{Method} & \textbf{CD$\downarrow$} & \textbf{AngOut\%$\downarrow$} & \textbf{NM\%$\downarrow$}\\
\midrule
Instant Meshes & 2.535 & 5.170 & 0.407\\
QuadriFlow & 5.207 & \textbf{2.410} & \underline{\textbf{0.000}}\\
QuadWild-BiMDF & \textbf{0.863} & 16.380 & \textbf{0.038}\\
SQuadGen & 3.240 & 19.540 & \underline{\textbf{0.000}}\\
\name{} (Ours) & \underline{\textbf{0.780}} & \underline{\textbf{1.170}} & \underline{\textbf{0.000}}\\
\bottomrule
\end{tabular}

\end{table}

\begin{figure}[!t]
 \centering
 \includegraphics[width=\columnwidth]{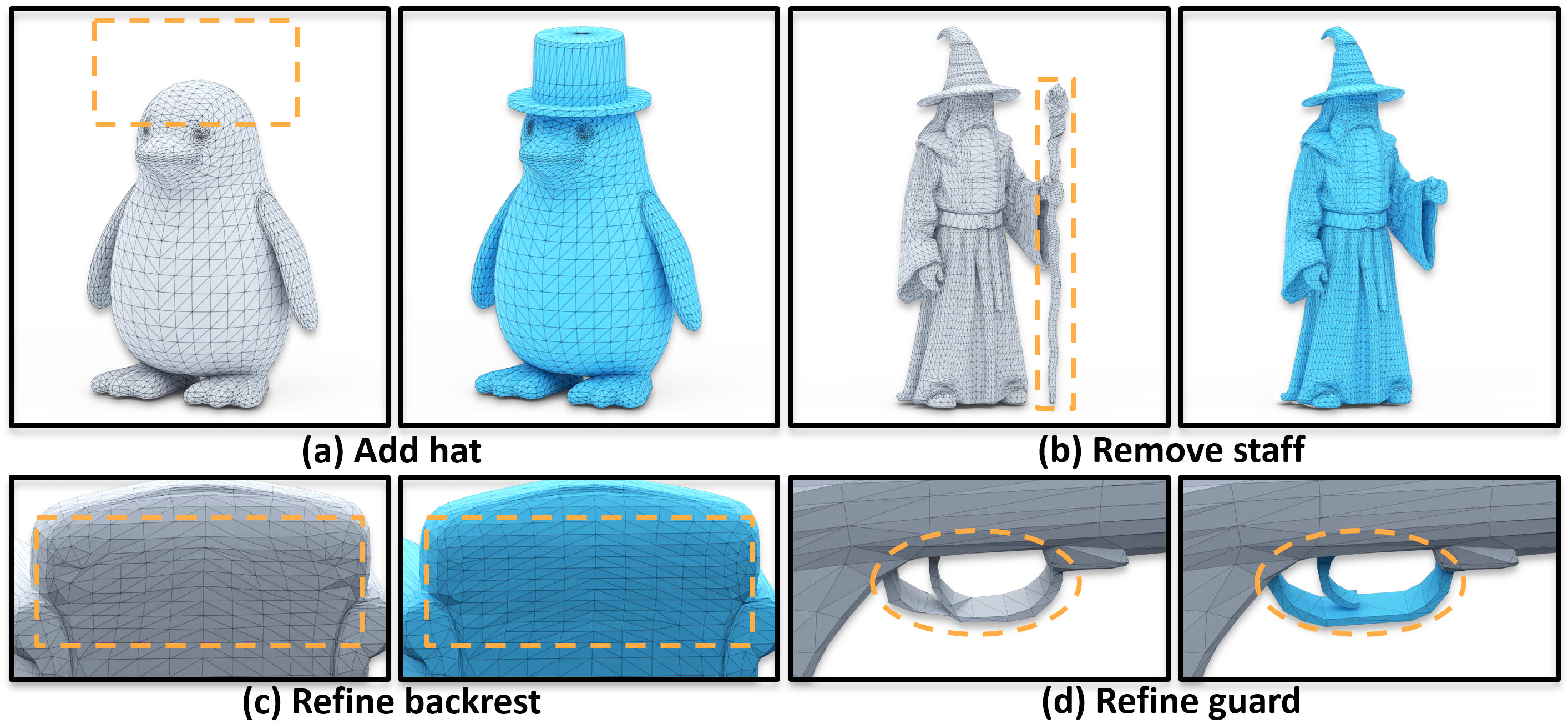}
 \vspace{-5mm}
 \caption{\textbf{Local editing and refinement.}
Inputs are shown in gray and results in blue for editing (top) and refinement (bottom).}
 \label{fig:edit-refine-results}
\end{figure}

Compared with the Regenerate + stitch baseline, which regenerates a whole mesh before cropping and stitching, regional shape editing achieves substantially stronger interface completeness alongside improved regional fidelity (\cref{tab:edit-compact} (a) and \cref{fig:edit-refine-results} (a-b)).
Throughout these local operations, vertices and faces strictly inside the protected region $K$ remain unchanged.
Additional edits on \name{}-generated meshes are shown in \cref{fig:supp-edit-gallery}.
Conditioned instead on the original surface, local refinement updates selected vertex positions and connectivity to repair local geometric and topological defects (\cref{fig:edit-refine-results} (c--d) and \cref{fig:refinement-cases}).
To quantitatively evaluate this capability, we construct $90$ local-refinement test cases. \name{} markedly reduces boundary defects and regional surface error while restoring clean reference connectivity (\cref{tab:edit-compact} (b); protocol details in \cref{sec:supp-local-editing}).

\begin{table}[!t]
\captionsetup{skip=3pt}
\caption{\textbf{Shape editing and refinement evaluation.}
Panel (a) evaluates local shape editing and (b) local refinement.
}
\label{tab:edit-compact}
\label{tab:auto-loop-refine}
\centering\experimenttable
\relax%
\setlength{\tabcolsep}{2pt}
\begin{minipage}[t]{0.49\columnwidth}
\centering
\resizebox{\linewidth}{!}{%
\begin{tabular}{l|ccc}
\toprule
\multicolumn{4}{c}{\textbf{(a) Local shape editing}}\\[-1pt]
\multicolumn{4}{c}{\textbf{106 tasks}}\\
\midrule
\textbf{Method} & \textbf{CD$\downarrow$} & \textbf{F1$\uparrow$} & \textbf{IC\%$\uparrow$}\\
\midrule
Regenerate + stitch & \textbf{0.437} & \textbf{0.884} & \textbf{7.759}\\
\name{} (Ours) & \underline{\textbf{0.357}} & \underline{\textbf{0.946}} & \underline{\textbf{99.313}}\\
\bottomrule
\end{tabular}}
\end{minipage}\hfill
\begin{minipage}[t]{0.49\columnwidth}
\centering
\resizebox{\linewidth}{!}{%
\begin{tabular}{l|ccc}
\toprule
\multicolumn{4}{c}{\textbf{(b) Local refinement}}\\[-1pt]
\multicolumn{4}{c}{\textbf{90 tasks}}\\
\midrule
\textbf{Stage} & \textbf{SL/1k$\downarrow$} & \textbf{CD$\downarrow$} & \textbf{Edge F1$\uparrow$}\\
\midrule
Before refinement & \textbf{5.846} & \textbf{1.163} & \textbf{0.498}\\
After refinement & \underline{\textbf{0.168}} & \underline{\textbf{0.754}} & \underline{\textbf{0.981}}\\
\bottomrule
\end{tabular}}
\end{minipage}
\end{table}

\subsection{User Study}
\label{sec:user-study}
We invited 40 professionals from the 3D industry for an anonymous subjective evaluation.
With method identities hidden, participants ranked triangle and quad outputs or selected a preferred local edit.
\Cref{tab:user-study} reports ranking scores and editing preferences; \name{} leads all three tasks with lower variance.
\Cref{sec:supp-user-study} details the cases, criteria, and aggregation.

\begin{table}[!t]
\centering\experimenttable
\caption{\textbf{User study of 40 complete responses.} Generation tasks report aggregated ranking scores and variances, while local editing reports first-choice preference rates.}
\label{tab:user-study}
\setlength{\tabcolsep}{2pt}
\begin{minipage}[t]{.43\columnwidth}\centering
\begin{tabular}{@{}lrr@{}}
\toprule
\multicolumn{3}{c}{\textit{Triangle generation}}\\
Method & Score$\uparrow$ & Var.\\
\midrule
BPT & 1.33 & 0.87\\
LATO.2 & 1.42 & 0.48\\
MeshFlow & 0.42 & 0.41\\
\name{} (Ours) & \textbf{\underline{2.83}} & 0.26\\
\bottomrule
\end{tabular}
\end{minipage}%
\hfill%
\begin{minipage}[t]{.55\columnwidth}\centering
\begin{tabular}{@{}lrr@{}}
\toprule
\multicolumn{3}{c}{\textit{Quad generation}}\\
Method & Score$\uparrow$ & Var.\\
\midrule
Instant Meshes & 1.33 & 0.38\\
QuadWild-BiMDF & 2.05 & 0.48\\
\name{} (Ours) & \textbf{\underline{2.62}} & 0.30\\
\midrule
\textit{Local editing} & \multicolumn{2}{r}{Preference (\%)$\uparrow$}\\
Regenerate + stitch & \multicolumn{2}{r}{26.4}\\
\name{} (Ours) & \multicolumn{2}{r}{\textbf{\underline{73.6}}}\\
\bottomrule
\end{tabular}
\end{minipage}%
\vspace{-4mm}
\end{table}

\FloatBarrier
\endgroup

\section{Conclusion}
\label{sec:conclusion}

We presented \name{}, a unified latent flow framework for generating, editing, and refining artist meshes.
By binding vertex latents to spatial anchors and evolving them under a regional asynchronous time field, \name{} makes local operations native to continuous mesh generation.
With a single triangle mesh model, our approach supports complete generation, explicit vertex budget control, and local editing and refinement while strictly preserving the protected region, whereas a separately trained variant sharing the same representation enables native quad generation.
Across evaluated benchmarks, \name{} delivers strong geometric fidelity and clean mesh topology, bridging generative synthesis with iterative modeling workflows.
We discuss limitations and future directions in \cref{sec:limitations} of the supplementary material.

\clearpage
\appendix
\maketitlesupplementary
\providecommand{\KC}[1]{#1}

\section{Evaluation Metrics and Conventions}
\label{sec:supp-metrics}

This section defines all metrics reported in the tables and figures. Unless a benchmark-specific protocol is noted otherwise, all meshes are evaluated within a shared normalized coordinate frame. Lower values indicate better performance for distance, defect, count-error, and runtime metrics, whereas higher values are better for agreement and coverage metrics.

\subsection{Common evaluation convention}

Given a generated mesh and its reference, let $\mathcal{G}=\{\mathbf{x}_i\}_{i=1}^{N_g}$ and $\mathcal{R}=\{\mathbf{y}_j\}_{j=1}^{N_r}$ denote area-weighted surface point samples with corresponding unit normals $\mathbf{n}_{\mathbf{x}_i}$ and $\mathbf{n}_{\mathbf{y}_j}$. The directed nearest-point operators are defined as
\begin{equation}
    \begin{aligned}
      \mathcal{P}_{g\!\to r}(\mathbf{x})&=\arg\min_{\mathbf{y}\in\mathcal{R}}\|\mathbf{x}-\mathbf{y}\|_2, \\
      \mathcal{P}_{r\!\to g}(\mathbf{y})&=\arg\min_{\mathbf{x}\in\mathcal{G}}\|\mathbf{x}-\mathbf{y}\|_2.
    \end{aligned}
  \label{eq:supp-nearest-point}
\end{equation}
For the triangle and native quad generation benchmarks, we sample $20{,}000$ points per surface after normalizing the bounding sphere to a diameter of $2.0$, whereas local editing and refinement evaluations use $8{,}192$ samples within the expanded evaluation region. All distance metrics in the tables are scaled by $10^2$ (e.g., a reported value of $0.710$ corresponds to an unscaled distance of $0.00710$).

For local operations, let $U_E$ and $U_R$ denote the expanded evaluation regions for local shape editing and defect refinement, respectively. For a mesh $M=(V,F)$ with vertex positions $\mathbf{p}_v$, we define $E_{U_R}(M)=\{\{u,v\}:\mathbf{p}_u,\mathbf{p}_v\in U_R\ \text{and }\{u,v\}\text{ is a mesh edge}\}$ as the set of regional edges within $U_R$.

\subsection{Surface fidelity metrics}

\paragraph{Chamfer distance (CD).}

We compute the symmetric, unsquared point-to-surface Chamfer distance by averaging the two directed nearest-point distances. Let $d_{g\!\to r,i} = \|\mathbf{x}_i - \mathcal{P}_{g\!\to r}(\mathbf{x}_i)\|_2$ and $d_{r\!\to g,j} = \|\mathbf{y}_j - \mathcal{P}_{r\!\to g}(\mathbf{y}_j)\|_2$ denote the nearest-point distances induced by Eq.~\eqref{eq:supp-nearest-point} for samples $\mathbf{x}_i$ and $\mathbf{y}_j$, respectively:
\begingroup
\small
\begin{equation}
  \mathrm{CD}(\mathcal{G},\mathcal{R})
  =\frac{1}{2N_g}\sum_{i=1}^{N_g}d_{g\!\to r,i}
  +\frac{1}{2N_r}\sum_{j=1}^{N_r}d_{r\!\to g,j}.
  \label{eq:supp-cd}
\end{equation}
\endgroup
For local editing and refinement (\cref{tab:edit-compact,tab:regional-ablation}), $\mathrm{CD}$ is evaluated by restricting both surfaces to the corresponding expanded local region ($U_E$ or $U_R$).

\paragraph{Hausdorff distance (HD95).}

To mitigate sensitivity to isolated sampling outliers, we report the 95th-percentile bidirectional Hausdorff distance. Let $Q_{0.95}$ denote the empirical 95th percentile:
\begingroup
\small
\begin{equation}
  \mathrm{HD95}(\mathcal{G},\mathcal{R})
  =\max\!\Big(
  Q_{0.95}\!\big(\{d_{g\!\to r,i}\}_{i=1}^{N_g}\big),\,
  Q_{0.95}\!\big(\{d_{r\!\to g,j}\}_{j=1}^{N_r}\big)
  \Big).
  \label{eq:supp-hd95}
\end{equation}
\endgroup

\paragraph{F-score (F1).}
Given a distance threshold $\tau=0.01$, precision and recall measure the fractions of generated and reference samples within $\tau$ of the opposite surface, respectively:
\begin{equation}
  \begin{aligned}
  \mathrm{Precision}&=\frac{1}{N_g}\sum_{i=1}^{N_g}\mathbf{1}\!\left[d_{g\!\to r,i}<\tau\right],\\
  \mathrm{Recall}&=\frac{1}{N_r}\sum_{j=1}^{N_r}\mathbf{1}\!\left[d_{r\!\to g,j}<\tau\right],\\
  \mathrm{F1}&=\frac{2\,\mathrm{Precision}\cdot\mathrm{Recall}}{\mathrm{Precision}+\mathrm{Recall}}.
  \end{aligned}
  \label{eq:supp-f1}
\end{equation}

For local shape editing (\cref{tab:edit-compact,tab:regional-ablation}), $\mathrm{F1}$ follows the same definition after restricting both surfaces to the expanded editing region $U_E$.

\paragraph{Normal consistency (NC).}
NC evaluates normal alignment across bidirectional nearest-point correspondences. We average the absolute dot products so that the score remains invariant to orientation flips:
\begin{equation}
  \begin{aligned}
  \mathrm{NC}(\mathcal{G},\mathcal{R})
  &=\frac{1}{2N_g}\sum_{i=1}^{N_g}\left|\mathbf{n}_{\mathbf{x}_i}\!\cdot\mathbf{n}_{\mathcal{P}_{g\!\to r}(\mathbf{x}_i)}\right|\\
  &\quad+\frac{1}{2N_r}\sum_{j=1}^{N_r}\left|\mathbf{n}_{\mathbf{y}_j}\!\cdot\mathbf{n}_{\mathcal{P}_{r\!\to g}(\mathbf{y}_j)}\right|.
  \end{aligned}
  \label{eq:supp-nc}
\end{equation}
NC is reported as a percentage in all tables.

\subsection{Mesh-structure metrics}

\paragraph{Non-manifold edges (NM\%).}
NM\% is the percentage of unique undirected mesh edges incident to more than two faces. For native quad evaluations, edges are counted directly on the polygonal output prior to any triangulation.

\paragraph{Valence irregularity (Val. irr.).}
A non-isolated vertex with valence $d$ is flagged as irregular if $|d-6|\geq 2$. Val.\ irr.\ denotes the fraction of irregular vertices among all non-isolated vertices.

\paragraph{SL/1k (small loops).}
We identify isolated connected components in the boundary-edge graph consisting of $3$--$8$ vertices where every vertex has a boundary degree of two, reported as the count per $1{,}000$ mesh faces. This metric captures small open boundary holes rather than regular interior edge loops.

\paragraph{Edge F1.} To evaluate local topological fidelity, we map generated vertices to clean reference vertex IDs via the recorded vertex correspondence, restrict both edge sets to $E_{U_R}$, and treat edges as undirected ID pairs. Let $E_g^R$ and $E_r^R$ denote the resulting generated and reference edge sets; from edge precision $P_e=|E_g^R\cap E_r^R|/|E_g^R|$ and recall $R_e=|E_g^R\cap E_r^R|/|E_r^R|$, we compute $\mathrm{Edge\mbox{-}F1}=2P_eR_e/(P_e+R_e)$. Whereas surface F1 measures geometric proximity, Edge-F1 evaluates exact regional connectivity agreement.

\paragraph{Quad angle outliers (AngOut\%).}
AngOut\% is computed per output as the percentage of the four corners of native quadrilateral faces whose angles fall outside $[30^\circ,150^\circ]$; residual triangles and other polygonal faces are excluded.

\paragraph{Zero-face (ZF) diagnostics.}
\textit{ZF meshes} denotes the number of evaluated meshes containing at least one degenerate triangle with an area below $10^{-12}$. \textit{ZF\%} is the mean fraction of faces per mesh satisfying this zero-area criterion, averaged across the evaluation set.

\subsection{Editing, refinement, and count-control metrics}

\paragraph{Interface completeness (IC\%).}
Prior to local editing, an expected interface edge is incident to exactly one preserved face in the protected region $K$ and one removed face in the editable region $U$, excluding pre-existing open boundaries. After assembly, the edge is deemed complete only if it remains incident to its original preserved face and is shared by exactly one newly generated face, free of duplicate or non-manifold incidences. IC\% is the percentage of expected interface edges that are complete.

\paragraph{Vertex-count error.}
Given a target vertex budget $B$ and the realized vertex count $V_{\mathrm{out}}$ (counting distinct vertices referenced by non-degenerate faces), the per-mesh relative error is $|V_{\mathrm{out}}-B|/B$. \textit{V-err.\%} is the mean relative error (in $\%$), \textit{P95\%} is the 95th-percentile relative error (in $\%$), and $S_{5\%}\%$ is the percentage of outputs achieving a relative error within $5\%$.

\paragraph{Runtime.}
\textit{Time} measures the end-to-end wall-clock runtime per input on a dedicated NVIDIA A800 GPU after warm-up, encompassing input preparation, conditioning, generation, decoding, and mesh export while excluding model loading.

\section{Method Details and Analysis}

\label{sec:supp_arch}

\label{sec:supp_training}

This section details data preprocessing, surface conditioning, mesh representation, flow training, and regional sampling.

\subsection{Training data and curation.}
We curate our training dataset from Objaverse-XL~\cite{deitke2023objaverse} and licensed artist assets.
Both the triangle and quad datasets undergo standard topological cleanup and strict de-duplication against all development and evaluation benchmarks, paired with topology-specific target construction and quality filtering. Within each variant, both the mesh VAE and the flow model are trained on its respective curated collection using ground-truth vertex positions and face connectivity.

\smallskip\noindent\textbf{(i) Triangle meshes:} To obtain clean triangle targets,
we weld coincident vertices,
repair non-manifold connectivity, split bow-tie singularities, and consistently orient face normals, discarding assets that remain topologically invalid. We retain meshes containing $100$ to $20{,}000$ vertices after cleanup with a maximum-to-median edge-length ratio no greater than $60$, yielding $532{,}699$ triangle meshes.

\smallskip\noindent\textbf{(ii) Quad meshes:} Since public 3D repositories frequently store artist quad meshes in triangulated form, we recover native quad-dominant targets by globally matching adjacent triangle pairs across shared manifold edges following QuadLink~\cite{quadlink}.
Candidate merges must be convex, with interior corner angles within $[25^\circ, 155^\circ]$ and a dihedral normal deviation below $30^\circ$. We maximize the number of valid pairings (each triangle matched at most once), breaking ties by minimal geometric distortion.
We first filter for candidates with $<1\%$ raw non-manifold edges, a median quad angle deviation $\le 20^\circ$, and at least $85\%$ surface coverage by recovered quads. Surviving meshes then undergo the same cleanup and topological validation steps described in~(i), enforcing zero tolerance for any residual non-manifoldness or topological defects to yield $278{,}512$ clean quad meshes.
\label{sec:supp-training-data}

\subsection{Spatial Anchors}
\label{sec:supp-anchor-capacity}

\paragraph{Anchor capacity.}
We discretize the anchor sequence capacity $L$ into seven preset levels $\mathcal{L} = \{1\text{K}, 2\text{K}, 4\text{K}, 8\text{K}, 10\text{K}, 15\text{K}, 20\text{K}\}$ via $L(B) = \min\{l \in \mathcal{L} : l \ge \min(\lceil 1.25B \rceil, 20\text{K})\}$, using the ground-truth vertex count during training and the requested vertex budget $B$ at inference.
Distinct target budgets can thus map to the same sequence length (for example, requests of $1{,}000$ and $1{,}500$ vertices both allocate $L = 2{,}000$ anchors).
Upward rounding provides anchor headroom (capped at $20\text{K}$) for spatially uneven vertex densities, while restricting $L$ to $\mathcal{L}$ enables efficient batched training with minimal padding overhead.

\paragraph{Vertex-to-anchor matching.}
\label{sec:supp-anchor-matching}

Given the candidate anchors, online training establishes a one-to-one correspondence between valid mesh vertices $V=\{v_j\}_{j=1}^{N}$ and anchors $A=\{a_i\}_{i=1}^{L}$ that minimizes Euclidean distance.
Specifically, we first compute all pairwise vertex-to-anchor distances and retain the $K_c=64$ nearest candidate anchors per vertex. We then execute $R=48$ synchronous proposal rounds.
In each round, every unmatched vertex proposes to its next available candidate, skipping anchors claimed in prior rounds. When an anchor receives multiple proposals, it accepts the vertex with the smallest Euclidean distance, breaking exact ties in favor of the lower vertex index.
Matched pairs are locked in and excluded from subsequent rounds, whereas rejected vertices advance to their next candidate. Following the $R=48$ rounds, any remaining unmatched vertices are sequentially assigned to their nearest unclaimed anchors.
This GPU-batched greedy algorithm efficiently approximates dense linear assignment without the overhead of an exact Hungarian solver.
Crucially, it guarantees the injective mapping required by the VAE: each anchor is matched to at most one vertex, each vertex is assigned at most once (and exactly once when $L\geq N$), and any surplus anchors remain unassigned.

\subsection{Mesh VAE}

\label{sec:supp-vae-training}

Our spatially addressable mesh VAE uses a hidden dimension of $768$ and $12$ attention heads across $10$ encoder layers
and $16$ decoder layers, totaling $532.45$M parameters.

\paragraph{Reconstruction Losses and Hyperparameters.}

\label{sec:supp-vae-heads}

Given $(z_i, a_i, c)$, the decoder supervises per-anchor occupancy probabilities $\hat o_i$ against binary labels $o_i$ via binary cross-entropy (BCE):
\begin{equation}
 \mathcal L_{\mathrm{occ}}
 =\mathbb E_i\!\left[\operatorname{BCE}(\hat o_i,o_i)\right].
 \label{eq:occupancy-loss}
\end{equation}
For each occupied anchor, the soft-label cross-entropy $\mathcal L_{\mathrm{coord}}$ in \cref{eq:expected-ce} supervises the predicted 512-bin coordinate distributions against a target truncated Gaussian centered at the ground-truth coordinate with a standard deviation of $0.5$ bins and a truncation radius of $2$ bins.
Simultaneously, the Gaussian posterior $q_\phi(z_i\mid M,c)$ of each occupied anchor's latent $z_i$ is regularized toward a standard normal prior via $\mathcal L_{\mathrm{KL}}=\mathbb E_i\!\left[D_{\mathrm{KL}}(q_\phi(z_i\mid M,c)\,\|\,\mathcal N(0,I))\right]$.
Combined with the balanced edge loss $\mathcal L_{\mathrm{edge}}$ (\cref{eq:edge-loss}) and orientation-invariant face loss $\mathcal L_{\mathrm{face}}$ (\cref{eq:face-loss}), we set the VAE loss weights in \cref{eq:vae-objective} to $\lambda_{\text{occ}} = 1.0$, $\lambda_{\text{coord}} = 1.0$, $\lambda_{\text{edge}} = 2.5$, $\lambda_{\text{face}} = 2.5$, and $\lambda_{\text{KL}} = 0.02$.

\paragraph{VAE training schedule.}
\label{sec:supp-vae-schedule}

We jointly train the triangle VAE encoder and decoder for $60{,}000$ steps, and initialize the quad VAE from the trained triangle VAE weights and fine-tune it for an additional $20{,}000$ steps. Both models are trained on $32$ NVIDIA A800 GPUs with a global batch size of $64$, using AdamW with weight decay $0.01$, BF16 computation, gradient clipping at norm $1.0$, and $81{,}920$ surface-conditioning points $c$ per sample.

\subsection{Flow Model and Training}
\label{sec:supp-time-field-details}

Building on our spatially addressable mesh VAE, the flow-matching transformer operates over $24$-dimensional anchor latents and comprises $34$ blocks with a hidden dimension of $1{,}728$, $18$ attention heads, and a $256$-dimensional sinusoidal time embedding, totaling $2.7$B parameters.
Prior to flow matching, the encoded latent means are standardized channel-wise using precomputed training statistics.

For each regional sample, we define the editable region $U$ by centering a box at a randomly selected target vertex and sizing it to contain approximately $10\%$, $25\%$, $50\%$, or $75\%$ of the target vertices with equal probability; the $15\%$ of anchors in $U$ closest to the protected set $K$ form the internal boundary band $R$, with regional time assignments $T_i$ and velocity supervision over $\Omega=\{i:T_i<1\}$ following Sec.~\ref{sec:local-support} (\cref{eq:async-loss}).

With the mesh VAE and surface encoder frozen, we train the flow transformer in two stages.
We first perform $40{,}000$ steps of synchronous whole-mesh pretraining, drawing $s\sim\mathcal U(0.05,0.95)$ and setting $T_i=s$ across all anchors.
Starting from this checkpoint, we train for an additional $12{,}000$ steps by sampling synchronous whole-mesh and regional asynchronous time fields with probabilities $0.25$ and $0.75$, respectively.

Both stages use $96$ NVIDIA A800 (80GB) GPUs with a global batch size of $96$, AdamW with weight decay $0.01$, gradient clipping at norm $1.0$, BF16 forward/backward computation with FP32 master weights, and an exponential moving average (EMA) decay of $0.99$.
In each stage, the learning rate independently warms up linearly to $10^{-5}$ over $500$ steps before cosine decay to $10^{-6}$.

We then initialize the quad flow model from the trained triangle flow weights and train it on the quad dataset for an additional $40{,}000$ steps, also using $96$ NVIDIA A800 (80GB) GPUs with a global batch size of $96$.

\begin{figure*}[t]
\centering
\includegraphics[width=\textwidth]{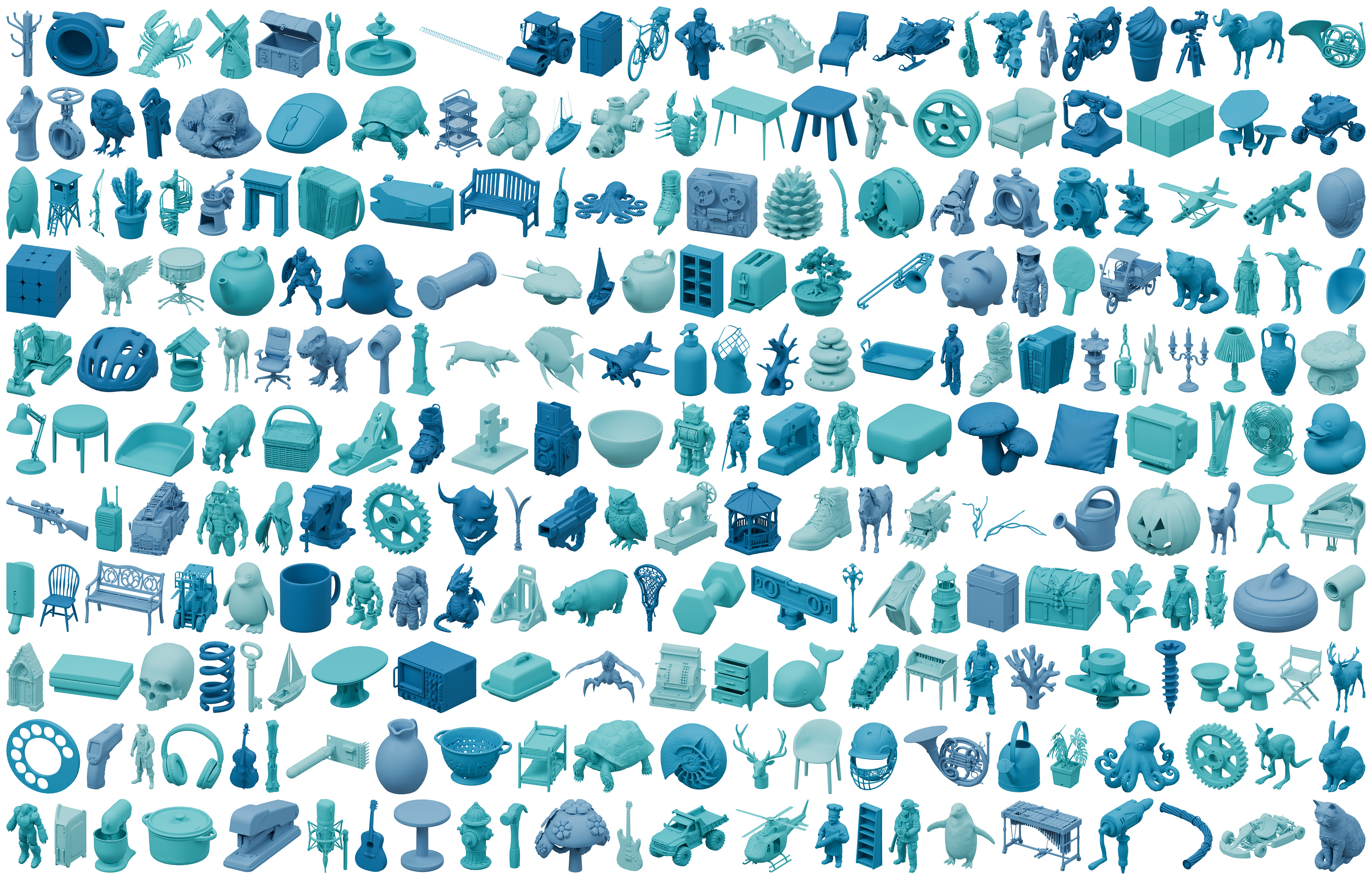}
\vspace{-8mm}
\caption{\textbf{Overview of the 250-surface generation test set collection.}
All 50 artist-created and 200 TRELLIS.2-generated input meshes are shown. The collection includes organic shapes and manufactured objects with varied silhouettes, part structures, and geometric detail.}
\label{fig:supp-generation-inputs}
\end{figure*}

\subsection{Inference Details}
\label{sec:supp-local-operation-details}

Our asynchronous flow formulation unifies complete-mesh generation, local shape editing, and local topology refinement under a single sampler over the anchor sequence $Y \in \mathbb{R}^{L \times 24}$.
Across all three tasks, we integrate from $t_0$ to $1$ using $16$ Heun steps (stochastic churn strength $50$) followed by a $6$-step Heun restart~\cite{xu2023restart}, with a classifier-free guidance (CFG) scale of $2.0$.
Governed by the regional time field $T_i(t)$ in \cref{eq:regional-flow}, the effective step rate $\alpha_i = \mathrm{d}T_i/\mathrm{d}t$ equals $1$ in the editable interior $U \setminus R$, $1/2$ in the boundary band $R$, and $0$ in the protected region $K$ (fixing $Y_K = Y_K^{\mathrm{ref}}$).
At each step $\Delta t = t_{k+1} - t_k$ conditioned on $(c^*, A^*, B^*)$, the Heun predictor--corrector update modulates the CFG-guided velocity $v_k = v_\theta(Y_k, T_k; c^*, A^*, B^*)$ by $\alpha$:
\begin{align}
Y^p &= Y_k + \Delta t \, (\alpha \odot v_k), \nonumber\\
Y_{k+1} &= Y_k + \frac{\Delta t}{2} \, \alpha \odot \big(v_k + v_\theta(Y^p, T_{k+1}; c^*, A^*, B^*)\big).
\label{eq:supp_heun}
\end{align}

The three operations share this exact solver and differ only in their region partition $(U, R, K)$ and starting time $t_0$.
Complete-mesh generation sets $U = \{1,\dots,L\}$ and $R = K = \emptyset$ ($\alpha_i = 1$ everywhere), initializing $Y_{t_0} \sim \mathcal{N}(0, I)$ at $t_0 = 0.05$.
Local editing and refinement instead partition anchors into $U$, $R \subset U$, and $K$, retaining clean reference latents on $K$ and assembling the decoded mesh per \cref{sec:local-support}: shape editing updates the target condition $c^*$ and resamples $U \setminus R$ from pure noise ($t_0 = 0$), whereas local refinement reuses $(c, Y)$ and partially noises $U \setminus R$ to $t_0 = 0.8$ (\cref{eq:refinement-start}).

\section{Evaluation Details}
\label{sec:supp_eval}

This section details the benchmark datasets, baseline configurations, and task-specific protocols used in our evaluations.

\subsection{Generation Input Collection}
\label{sec:supp-generation-inputs}

\Cref{fig:supp-generation-inputs} presents the full benchmark collection of $250$ distinct input surfaces, comprising $50$ artist-created meshes and $200$ meshes generated via TRELLIS.2~\cite{trellis2}. The benchmark spans everyday objects, mechanical parts, vehicles, architecture, plants, animals, and characters, covering both organic and hard-surface geometry with slender parts, openings, articulated structures, and curved surfaces.

Across both the triangle and quad generation benchmarks, methods supporting explicit vertex-count conditioning are queried with a target budget of $2{,}000$ vertices for TRELLIS.2 inputs and the ground-truth vertex count for artist-mesh inputs (capped at each method's maximum supported capacity; see Tab.~\ref{tab:supp_capacity}). Methods without explicit count control are evaluated under their default capacity settings.

\subsection{Image-conditioned Comparison}
\label{sec:supp-image-comparison}
For the image-conditioned comparisons in \cref{fig:image-conditioned-comparison}, Commercial A and Commercial B take each input image directly, whereas we use TRELLIS.2~\cite{trellis2} to convert the image into an intermediate conditioning surface before synthesizing the final artist mesh with \name{}.
Commercial A, Commercial B, and \name{} are all queried with a requested budget of 3,000 vertices.

\subsection{Triangle Mesh Generation}

\label{sec:supp-main-generation-details}

\paragraph{Baseline configurations and capacity limits.}
In \cref{tab:unified-generation} of the main paper, all baselines are evaluated using their official implementations with native post-processing enabled, whereas \name{} is evaluated directly on raw outputs without post-processing. Their declared capacity limits and count-control capabilities are summarized in Tab.~\ref{tab:supp_capacity}.

\begin{table}[!t]
\centering\experimenttable
\caption{\textbf{Reported and configured capacity limits.}
Capacity is given in each method's native face or vertex units. Explicit target indicates support for inference-time vertex-budget conditioning rather than an exact count guarantee.
}
\label{tab:supp_capacity}
\resizebox{\columnwidth}{!}{%
\begin{tabular}{l|cc|l}
\toprule
\textbf{Method} & \textbf{Declared limit} & \textbf{Explicit target} & \textbf{Unit / Scope} \\
\midrule
FastMesh~\cite{fastmesh} & $4{,}000$ & No & Vertices (released V4K cap) \\
BPT~\cite{scaling_mesh_generation} & $8{,}000$ & No & Faces (compressed token limit) \\
TreeMeshGPT~\cite{treemeshgpt} & $5{,}500 / 11{,}000$ & No & Faces (7-bit / 9-bit limits) \\
DeepMesh~\cite{deepmesh} & $30{,}000$ & No & Faces (reported training cap) \\
MeshAnything V2~\cite{meshanythingv2} & $1{,}600$ & No & Triangles (configured limit) \\
MeshFlow~\cite{meshflow} & $4{,}096$ & Yes & Vertices (configured slot limit) \\
MeshRipple (10k)~\cite{meshripple} & $10{,}000$ & No & Faces (full-attention limit) \\
MeshRipple (20k)~\cite{meshripple} & $20{,}000$ & No & Faces (native sparse attention) \\
LATO.2~\cite{lato2} & $10{,}000$ & Yes & Vertices (topology-stage limit) \\
\textbf{\name{} (Ours)} & $\mathbf{20{,}000}$ & \textbf{Yes} & \textbf{Vertices (configured limit)} \\
\bottomrule
\end{tabular}%
}
\end{table}

\subsection{Native Quad Generation}

In \cref{tab:quad} of the main paper (Sec.~\ref{sec:quad-results}), we evaluate native quad generation across the $250$ test inputs using $\text{CD}$, $\text{NM\%}$, and $\text{AngOut\%}$. To match target complexity for classical quad remeshers, let $Q$ and $T$ denote the realized quadrilateral and triangle counts of the corresponding \name{} output, yielding a triangle-equivalent face count $B_{\text{eq}} = 2Q + T$ and a target quad budget $Q_d = \max(1,\lceil B_{\text{eq}} / 2 \rceil)$. QuadriFlow~\cite{quadriflow2018} is configured with a target face count of $Q_d$ (though its realized count may deviate). Instant Meshes~\cite{instantmeshes2015} is run with fourfold rotational and positional symmetry, a $40^\circ$ crease angle, boundary alignment, two smoothing iterations, and a coarse face target of $\max(1, \lfloor(Q_d + 2)/4\rfloor)$ to account for four-way subdivision during extraction. QuadWild-BiMDF~\cite{quadwild2021} uses its default $35^\circ$ sharp-feature angle with its edge-scale parameter set to $\sqrt{A / Q_d} / \bar{l}$, where $A$ and $\bar{l}$ denote the input surface area and mean edge length, respectively. SQuadGen~\cite{squadgen2026}, which does not support explicit face-count targeting, is evaluated using its released weights at a latent resolution of $8{,}192$, default isotropic remeshing, and quad extraction with a target edge length of $0.2\%$ of the bounding-box diagonal and a $10^\circ$ feature angle.

\subsection{Vertex-Count Control}

\label{sec:supp-count-control}

\begin{figure}[!t]
\centering
\includegraphics[width=\columnwidth]{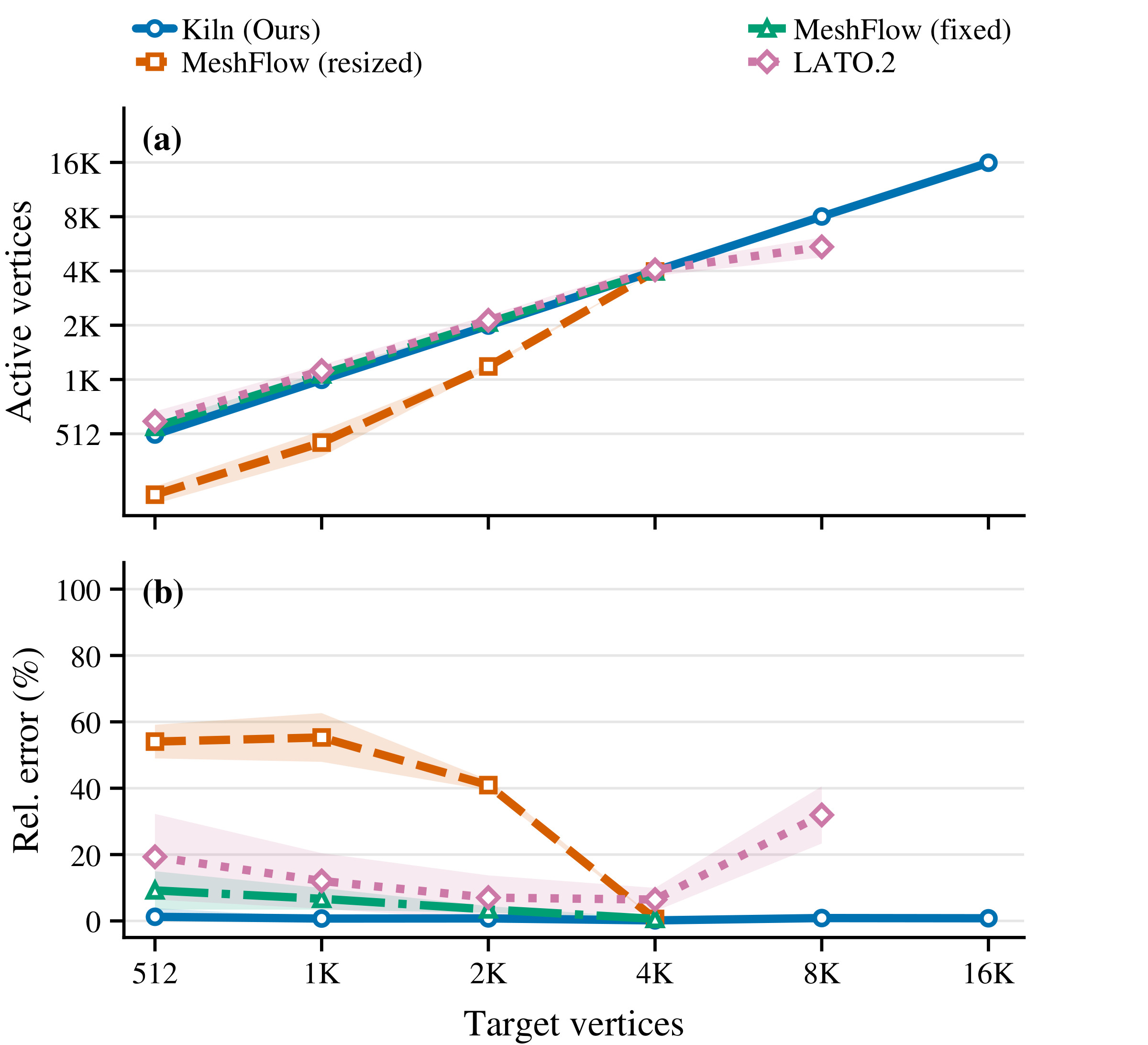}
\vspace{-6mm}
\caption{\textbf{Requested versus realized vertex counts.}
Top: realized active vertices versus requested budgets, with both axes shown on a base-2 logarithmic scale.
Bottom: absolute relative vertex-count error ($\%$).
Budgets are evaluated up to each method's supported capacity: $512$--$4{,}096$ vertices for the fixed-slot and resized-slot configurations of MeshFlow~\cite{meshflow}, $512$--$8{,}192$ for LATO.2~\cite{lato2}, and $512$--$16{,}384$ for \name{}.}
\label{fig:supp_count_curve}
\end{figure}

To evaluate explicit vertex-count control (\cref{tab:vertex-control} of the main paper), we compare \name{} against MeshFlow~\cite{meshflow} and LATO.2~\cite{lato2}, the two evaluated baselines that support inference-time vertex-budget conditioning.
For MeshFlow, we consider two variants using either a fixed $4{,}096$-slot sequence or a sequence length scaled proportionally to the requested budget.
Across all $250$ test inputs, we query power-of-two target budgets starting from $512$ vertices and evaluate each method up to its supported capacity: $4{,}096$ vertices for both MeshFlow variants, $8{,}192$ for LATO.2, and $16{,}384$ for \name{}.
\Cref{fig:supp_count_curve} plots budget-tracking accuracy across these ranges, while \cref{fig:vertex-count-qualitative} illustrates how \name{} generates progressively denser, well-structured tessellations up to $20\text{K}$ vertices from a single input surface.

\subsection{Local Editing and Refinement}

\label{sec:supp-local-editing}

\paragraph{Benchmark construction and protocol.}
We evaluate local mesh operations across $196$ tasks, comprising $106$ local shape edits and $90$ local mesh refinements (\cref{tab:edit-compact} of the main paper). 
For shape editing, we manually curate source--target pairs, each consisting of a source mesh, a locally modified target surface $c^*$, and a corresponding editable region $U$.
For mesh refinement, we construct $90$ test cases by synthetically injecting localized geometric and topological defects into clean reference meshes, including boundary holes, corrupted faces, and perturbed vertices with irregular edge flows.

\paragraph{Local shape editing.}

In \cref{tab:edit-compact} (a), we compare \name{} against a \textit{Regenerate + stitch} baseline across the $106$ shape-editing tasks. \textit{Regenerate + stitch} first runs whole-mesh generation conditioned on the modified full surface $c^*$ using the same target vertex budget. It then crops the generated mesh to the editable region $U$, retains the original mesh inside the protected region $K$, and stitches the two parts along the interface by snapping and welding boundary vertices. Because independent whole-mesh generation does not preserve the vertex layout or edge flow along $\partial K$, post-hoc cropping and stitching frequently introduces seam gaps, duplicate faces, or non-manifold incidences. By contrast, \name{} directly synthesizes the selected region within the surrounding context of $K$, conforming to $c^*$ while maintaining seamless interface connectivity. \Cref{fig:edit-refine-results} (a,b) shows qualitative examples of adding a top hat and removing a staff while leaving the unedited mesh strictly intact.

\begin{figure}[!t]
  \centering
  \includegraphics[width=\columnwidth]{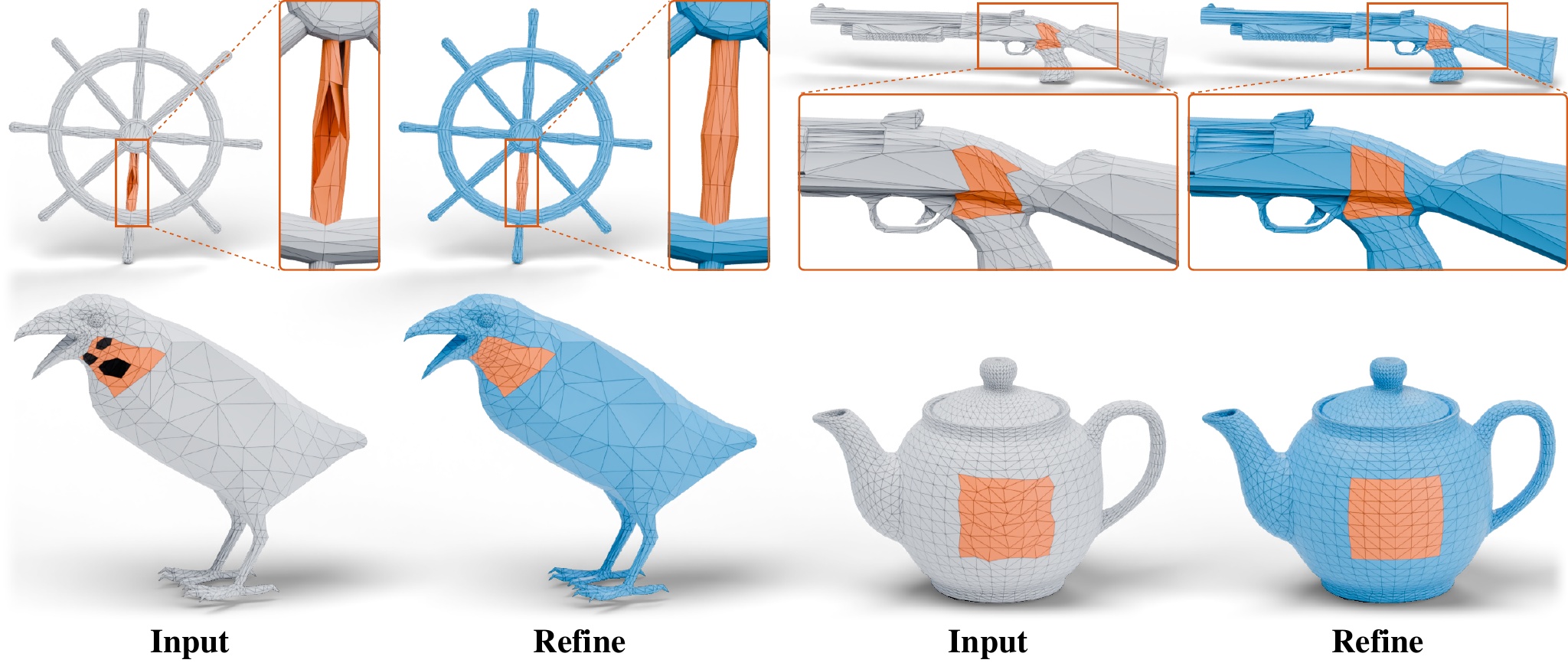}
  \vspace{-4mm}
  \caption{\textbf{Additional local refinement results.} 
Orange highlighted faces indicate the selected local support. Close-up insets compare the local mesh topology before and after refinement.}
  \label{fig:refinement-cases}
\end{figure}

\paragraph{Local mesh refinement.}

Conditioned on the original surface $c$, our refinement mode jointly updates local vertex positions and face connectivity within the target region $U_R$ while strictly preserving the surrounding mesh $K$. In \cref{tab:edit-compact} (b), we conduct a paired before-and-after evaluation across the $90$ local refinement tasks. Starting from the corrupted inputs (\textit{Before refinement}), local refinement (\textit{After refinement}) closes artificial boundary holes ($\text{SL/1k}$), reduces regional surface error ($\text{CD}$), and restores clean mesh connectivity that closely matches the ground-truth topology ($\text{Edge F1}$). Qualitative examples on a chair backrest and a trigger guard are shown in \cref{fig:edit-refine-results} (c,d), with additional cases provided in \cref{fig:refinement-cases}.

\section{Component Ablations}
\label{sec:supp-ablation}
\label{sec:experiment-ablations}

\begin{figure}[!t]
  \centering
  \includegraphics[width=\columnwidth]{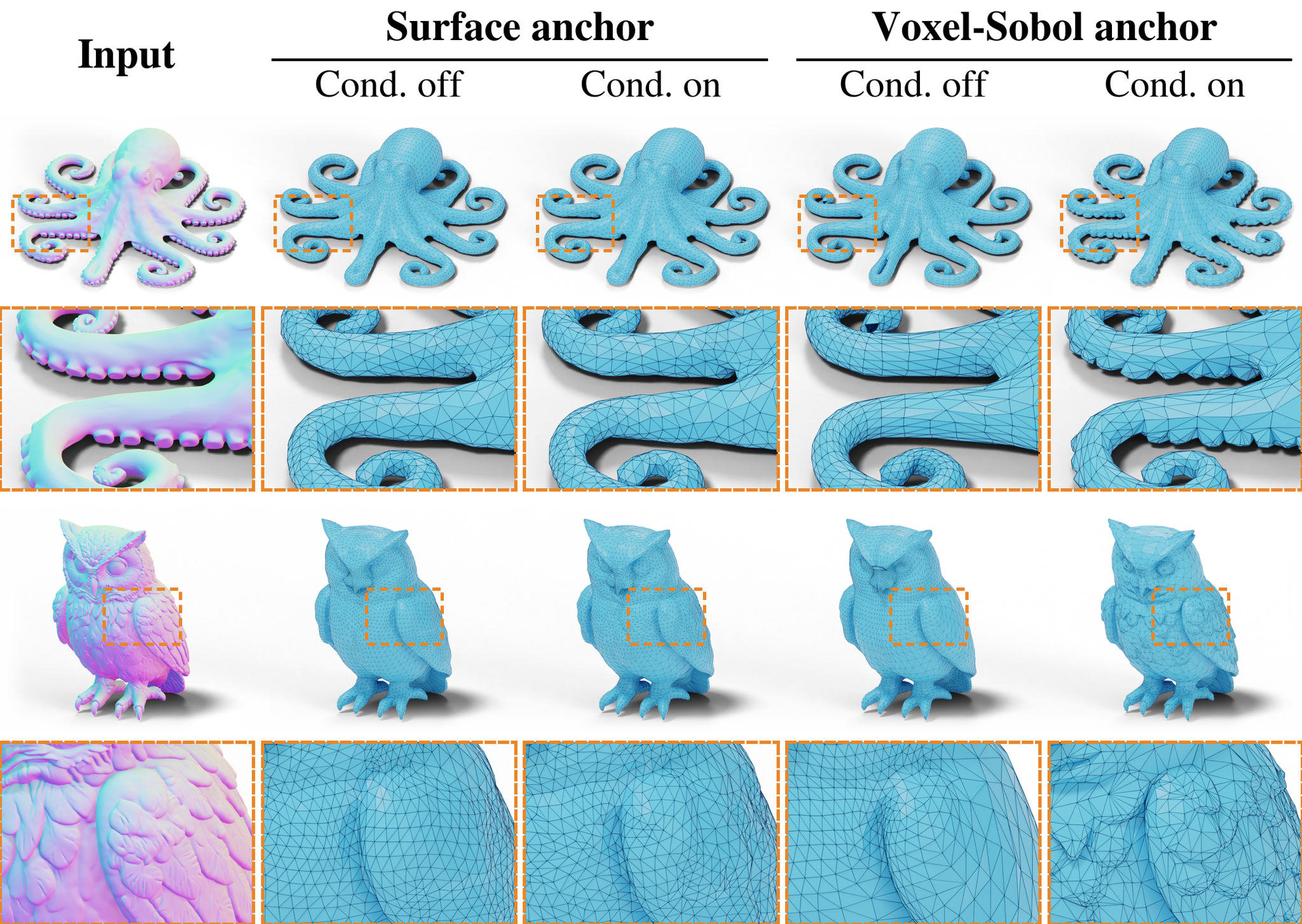}
  \vspace{-4mm}
  \caption{\textbf{Ablation on anchor placement and flow conditioning.} 
    Qualitative comparison between surface-sampled and Voxel--Sobol anchors, evaluated with and without surface conditioning.}
  \label{fig:anchor-condition-qualitative}
\end{figure}

\subsection{Surface conditioning beyond anchors.}
\label{sec:anchor-condition-diagnostic}

We first examine how anchor placement governs the model's reliance on the dense surface condition $c$.
Because sparse anchors cannot capture fine geometry, detailed structure should be drawn from $c$ rather than leaked through surface-sampled anchors acting as a shape shortcut.
To isolate the effect of anchor placement, we compare two independently trained VAE--flow pipelines sharing the exact same training schedule: \name{}'s default pipeline using coarse voxel anchors, and a surface-anchor variant that samples anchors directly from the proxy surface $H$ and retrains both the VAE and the flow model accordingly. We evaluate both pipelines across all 250 test inputs (50 artist-created meshes and 200 TRELLIS.2-generated meshes; \cref{tab:flow-surface-conditioning}). At inference, we toggle the surface condition $c$ while holding the anchors, initial noise, and requested vertex budget fixed within each pipeline: ``On'' supplies the encoded surface features $c$ to both the flow model and the VAE decoder, whereas ``Off'' replaces $c$ with all-zero null tokens across both modules. With $c$ enabled, coarse anchors consistently outperform surface-sampled anchors across all metrics. Crucially, disabling $c$ causes a sharp degradation for coarse anchors (F1 drops from $0.743$ to $0.538$, and CD rises from $0.782$ to $1.233$), whereas surface anchors exhibit only a minor decline (F1 drops from $0.705$ to $0.678$). This contrast confirms that surface-sampled anchors induce shortcut learning, while coarse voxel anchors compel the pipeline to actively exploit the dense surface condition $c$ for fine geometric reconstruction.
Qualitative comparisons in \cref{fig:anchor-condition-qualitative} visually corroborate this behavior.

\subsection{Coordinate Decoding}
\label{sec:coordinate-decoding}

\begin{figure}[!t]
 \centering
 \includegraphics[width=\columnwidth]{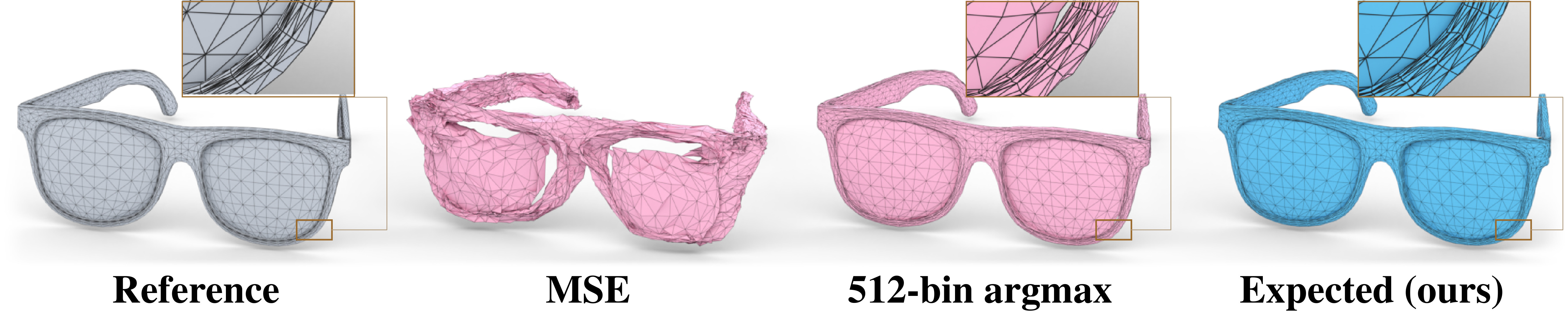}
 \vspace{-8mm}
 \caption{\textbf{VAE coordinate-decoding ablation.}
 We compare VAE coordinate reconstructions across three separately trained models: (i) direct MSE coordinate regression, (ii) $512$-bin classification trained with one-hot cross-entropy (CE) and discrete argmax readout, and (iii) our $512$-bin distribution trained with soft-label CE and continuous expected-bin readout. Insets highlight quantization and readout artifacts along the curved rim.}
 \label{fig:ce-mse-reconstruction}
\end{figure}

To evaluate the impact of coordinate supervision and decoding parameterization on geometric fidelity and face degeneracy, we train three separate VAE models under the standard $60{,}000$-step schedule (Sec.~\ref{sec:supp-vae-schedule}):
(i) direct MSE coordinate regression, (ii) $512$-bin classification trained with one-hot cross-entropy (CE) and discrete argmax readout, and (iii) our $512$-bin distribution trained with soft-label CE and continuous expected-bin readout (\cref{eq:expected-ce}).
We evaluate reconstructions across $128$ development meshes from both clean latent means ($\sigma = 0$) and Gaussian-perturbed latents ($\sigma = 0.3$) in \cref{fig:ce-mse-reconstruction,tab:decoder-condition-ablations}.
Under both regimes, $512$-bin argmax quantization snaps proximate vertices to identical discrete bins, introducing zero-area faces on $113/128$ clean meshes ($\text{ZF\%} = 3.741\%$) and $115/128$ noisy meshes ($\text{ZF\%} = 3.525\%$) despite low global $\text{CD}$.
Conversely, our continuous expected-bin readout preserves sub-bin precision, achieving the lowest $\text{CD}$ ($0.368$ clean, $0.371$ noisy) while matching MSE regression with zero degenerate faces ($0/128$ ZF meshes, $\text{ZF\%} = 0.000\%$) across both conditions.

\begin{table}[!t]
\caption{\textbf{Surface conditioning.} We compare Kiln's default coarse-anchor VAE--flow pipeline against an independently trained surface-anchor variant under the same training schedule. ``On'' supplies the encoded surface condition $c$ to both the flow model and the VAE decoder, whereas ``Off'' replaces $c$ with all-zero null tokens in both modules. Metric definitions and units are given in Sec.~\ref{sec:supp-metrics}.}
\label{tab:flow-surface-conditioning}
\centering\experimenttable
\setlength{\tabcolsep}{3pt}
\renewcommand{\arraystretch}{0.95}
\begin{tabular}{lc|ccc}
\toprule
\textbf{Anchors} & \textbf{Condition $c$} & \textbf{CD$\downarrow$} & \textbf{F1$\uparrow$} & \textbf{NC\%$\uparrow$}\\
\midrule
Coarse & On & \underline{\textbf{0.782}} & \underline{\textbf{0.743}} & \underline{\textbf{92.398}}\\
Coarse & Off & 1.233 & 0.538 & 83.708\\
Surface & On & \textbf{0.858} & \textbf{0.705} & \textbf{91.060}\\
Surface & Off & 0.919 & 0.678 & 89.679\\
\bottomrule
\end{tabular}
\end{table}

\begin{table}[!t]
\caption{\textbf{Coordinate decoding under clean and noisy latents.} We compare coordinate readout methods on the same 128 meshes. Metric definitions and units are given in Sec.~\ref{sec:supp-metrics}.}
\label{tab:decoder-condition-ablations}
\label{tab:supp_readout_noise}
\centering\experimenttable
\setlength{\tabcolsep}{2.5pt}
\renewcommand{\arraystretch}{0.95}
\begin{tabular}{l|ccc}
\toprule
& \multicolumn{3}{c}{\shortstack{Clean latents\\($\sigma=0$)}}\\
\cmidrule(lr){2-4}
\textbf{Readout} & \textbf{CD$\downarrow$} & \textbf{ZF meshes$\downarrow$} & \textbf{ZF\%$\downarrow$}\\
\midrule
MSE & 1.238 & \underline{\textbf{0/128}} & \underline{\textbf{0.000}}\\
CE argmax & \textbf{0.370} & 113/128 & 3.741\\
CE expectation & \underline{\textbf{0.368}} & \underline{\textbf{0/128}} & \underline{\textbf{0.000}}\\
\bottomrule
\end{tabular}
\par\smallskip
\begin{tabular}{l|ccc}
\toprule
& \multicolumn{3}{c}{\shortstack{Noisy latents\\($\sigma=0.3$)}}\\
\cmidrule(lr){2-4}
\textbf{Readout} & \textbf{CD$\downarrow$} & \textbf{ZF meshes$\downarrow$} & \textbf{ZF\%$\downarrow$}\\
\midrule
MSE & 1.222 & \underline{\textbf{0/128}} & \underline{\textbf{0.000}}\\
CE argmax & \textbf{0.387} & 115/128 & 3.525\\
CE expectation & \underline{\textbf{0.371}} & \underline{\textbf{0/128}} & \underline{\textbf{0.000}}\\
\bottomrule
\end{tabular}
\end{table}

\begin{table}[!t]
\caption{\textbf{Regional training ablation.} All four variants receive $12{,}000$ additional updates from the same whole-mesh-pretrained flow checkpoint.}
\label{tab:regional-ablation}
 \centering\experimenttable
 \setlength{\tabcolsep}{5pt}
 \renewcommand{\arraystretch}{0.95}
 \begin{tabular}{l|ccc}
 \toprule
 \textbf{Variant} & \textbf{CD$\downarrow$} & \textbf{F1$\uparrow$} & \textbf{IC\%$\uparrow$}\\
 \midrule
 Whole-mesh synchronous & \textbf{0.496} & 0.834 & 80.184\\
Regional loss only & 0.597 & 0.721 & 91.084\\
Regional async., no band & 0.544 & \textbf{0.852} & \textbf{96.139}\\
\midrule
Full recipe (Ours) & \underline{\textbf{0.357}} & \underline{\textbf{0.946}} & \underline{\textbf{99.313}}\\
 \bottomrule
 \end{tabular}
\end{table}

\subsection{Regional Training}
To evaluate how regional supervision, asynchronous time conditioning, and the internal boundary band contribute to local editing, we compare four training configurations across the $106$ shape-editing tasks (\cref{tab:regional-ablation}). All variants are initialized from the same synchronously pretrained whole-mesh checkpoint and fine-tuned for $12{,}000$ steps under identical optimization settings (Sec.~\ref{sec:supp-time-field-details}).

\smallskip\noindent\textbf{(i) Whole-mesh synchronous.}
We continue synchronous whole-mesh training ($T_i = s$ across all anchors) and supervise velocity over the entire mesh, introducing regional sampling only at inference. This configuration yields the lowest interface completeness ($\text{IC\%} = 80.184\%$), indicating that inference-time masking alone cannot reliably stitch edited regions to the preserved boundary.

\smallskip\noindent\textbf{(ii) Regional loss only.}
We restrict velocity supervision to the editable anchors ($\Omega = U$) while retaining a shared synchronous time $T_i = s$ across all anchors. Because the protected region $K$ is still corrupted during training rather than provided as clean context, this variant improves seam closure ($\text{IC\%} = 91.084\%$) relative to whole-mesh training but degrades geometric fidelity ($\text{CD} = 0.597$, $\text{F1} = 0.721$).

\smallskip\noindent\textbf{(iii) Regional async., no band.}
We combine regional supervision with binary asynchronous time conditioning, keeping protected anchors clean ($T_K = 1$) while assigning a uniform noise level ($T_U = s$) to all editable anchors without an intermediate boundary band. Providing clean surrounding context substantially recovers geometric fidelity and boosts interface completeness to $96.139\%$.

\smallskip\noindent\textbf{Full recipe (Ours).}
Our full recipe additionally introduces the internal boundary band $R \subset U$ at an intermediate time $T_R = (1+s)/2$, creating a smooth transition between the noisy editable interior $U \setminus R$ and the clean protected region $K$. Combining all three components achieves the best geometric fidelity ($\text{CD} = 0.357$, $\text{F1} = 0.946$) and highest interface completeness ($\text{IC\%} = 99.313\%$).

\begingroup
\section{User Study Protocol}
\label{sec:supp-user-study}
\paragraph{Study design and interface.}
We conduct a blind perceptual study across $25$ representative test cases: $10$ triangle-mesh generation comparisons (\name{}, BPT~\cite{scaling_mesh_generation}, LATO.2~\cite{lato2}, and MeshFlow~\cite{meshflow}), $6$ native quad-mesh generation comparisons (\name{}, Instant Meshes~\cite{instantmeshes2015}, and QuadWild-BiMDF~\cite{quadwild2021}), and $9$ local shape-editing comparisons (\name{} versus the \textit{Regenerate + stitch} baseline).
All methods are evaluated using outputs from their standard inference pipelines under their respective capacity settings (Sec.~\ref{sec:supp_eval}).
For each question, an interactive web interface presents the input reference alongside all candidate outputs in synchronized 3D viewports, allowing participants to freely rotate, pan, and zoom the camera, toggle native wireframe overlays, and inspect a fixed-view reference rendering. Local editing tasks additionally display the modified target geometry and a brief text prompt describing the intended edit.
To ensure blind evaluation, method identities are anonymized using letter labels whose assignment is permuted across questions.

\paragraph{Participants and evaluation criteria.}
We recruited 40 professional 3D artists and practitioners for the study; all responses were collected anonymously without recording personally identifiable information, and only fully completed questionnaires were retained for analysis.
Participants were instructed to holistically evaluate each candidate based on task-specific geometric and topological criteria:
(i) for \textit{triangle generation}, silhouette accuracy, overall proportions, preservation of fine geometric details, tessellation regularity, and absence of topological defects (e.g., missing surfaces, disconnected fragments, or jagged boundaries);
(ii) for \textit{quad generation}, edge-flow alignment, face regularity, and preservation of thin structures; and
(iii) for \textit{local editing}, fidelity to the target shape, seam continuity across the edit interface, and strict preservation of the unedited region.

\paragraph{Scoring and aggregation.}
For the multi-way generation tasks, participants provide a strict overall ranking without ties (selecting the top three of four candidates for triangle generation and the top two of three for quad generation, with the remaining candidate assigned the final rank).
Ranks $1$, $2$, and $3$ are awarded $3$, $2$, and $1$ points, respectively, and rank $4$ receives $0$ points; consequently, composite scores range from $0$ to $3$ for triangle generation and from $1$ to $3$ for quad generation, and should be compared strictly within each task.
For the pairwise local editing tasks, participants simply select their preferred output.
We report the mean rank score and descriptive variance (divisor $N$) for the triangle and quad generation tasks, and the preference rate (the percentage of pairwise judgments in which a method is preferred) for pairwise local-editing tasks.
\endgroup

\begin{figure}[!tbp]
\centering
\KC{\includegraphics[width=\linewidth]{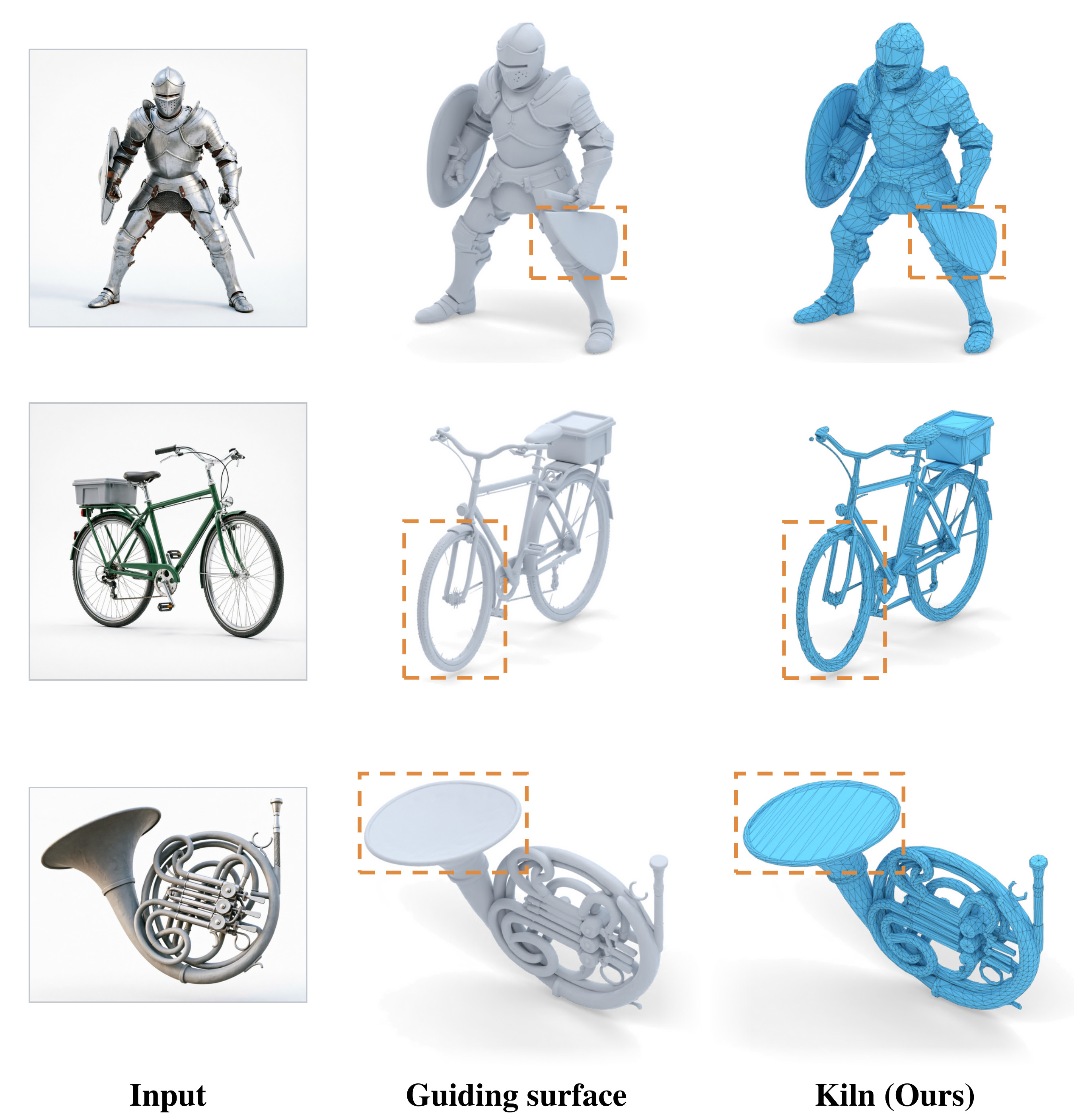}}
\vspace{-6mm}
\caption{\textbf{Inherited surface artifacts.} Errors in guiding surfaces can persist in Kiln outputs.}
\vspace{-2mm}
\label{fig:upstream-limitations}
\end{figure}

\section{Limitations and Future Work}
\label{sec:limitations}

While \name{} demonstrates strong performance across whole-shape generation, local editing, and mesh refinement, several limitations remain. First, because our framework is surface-conditioned, output quality is inherently bounded by the upstream guiding geometry. As shown in \cref{fig:upstream-limitations}, when upstream 3D shape generators or image-to-3D reconstructions introduce geometric artifacts or floaters, \name{} tends to inherit these defects rather than semantically removing them, leading to degraded outputs. Second, although \name{} achieves zero non-manifold edges on our generation benchmarks under the fixed output pipeline, manifoldness remains an empirical property rather than a strict architectural guarantee. Moreover, while our structural metrics confirm local regularity in vertex valence and quad angles, pure surface conditioning does not enforce globally meaningful edge flow, such as joint-aligned loops required for rigging and animation deformation.

Beyond topological and semantic properties, our current representation encodes purely static geometry without modeling part articulation or temporal dynamics. Finally, despite supporting a relatively large vertex budget among existing artist-mesh generators, \name{} is bounded by a hard sequence cap of $20\text{K}$ vertices, which restricts direct single-pass synthesis of ultra-dense production assets or full-scale scenes. Improving robustness against imperfect conditioning surfaces, incorporating deformation-aware priors, extending our latent flow to articulated or dynamic meshes, and scaling beyond the $20\text{K}$-vertex ceiling represent promising directions for future work.

\section{Additional Qualitative Results}

\label{sec:supp_qualitative}

\begin{figure*}[p]
\centering
\includegraphics[width=\textwidth,height=\textheight,keepaspectratio]{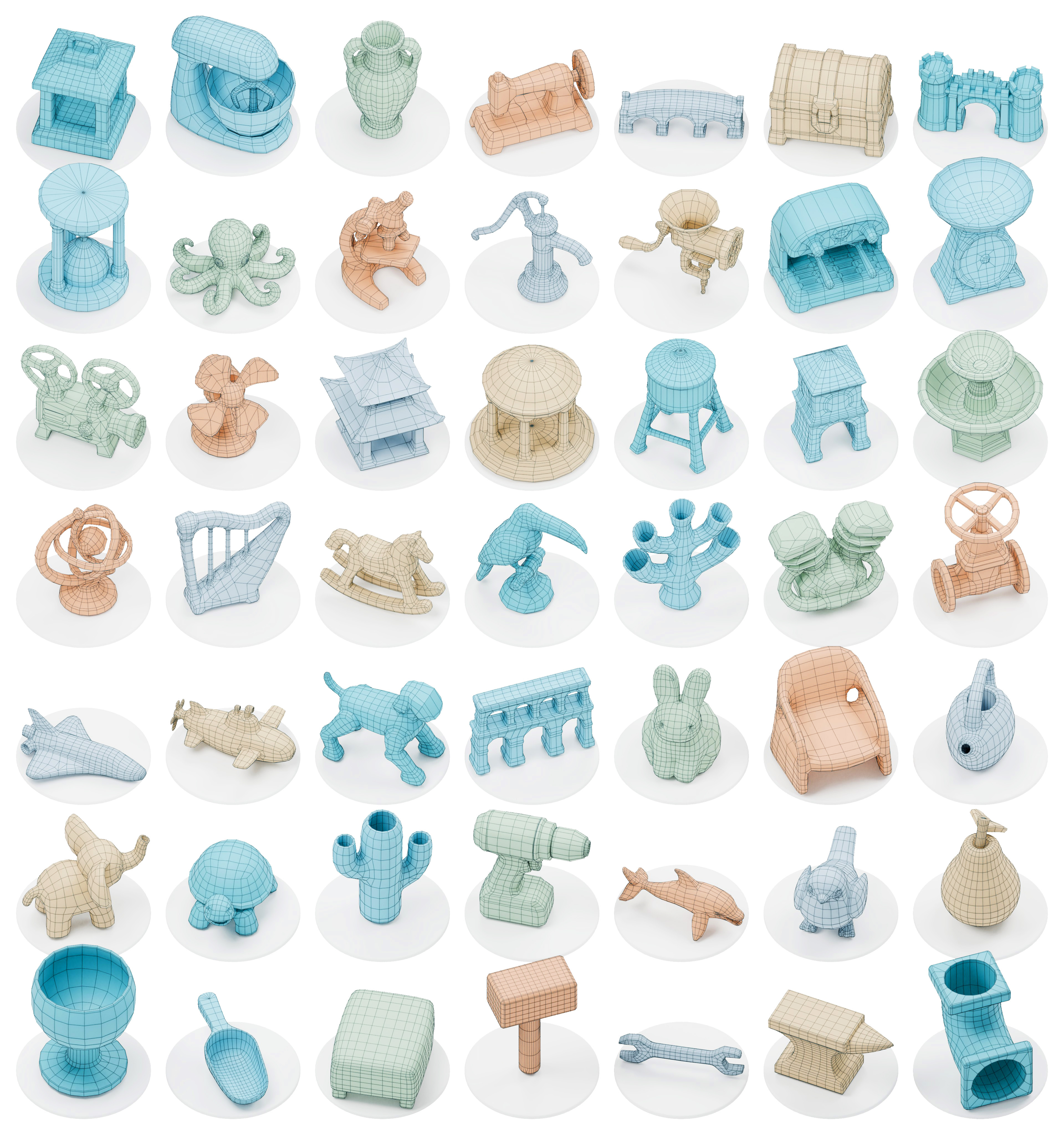}
\caption{\textbf{Quad generation gallery.}
Results from \name{} across a variety of shapes, with native polygon edges shown.}
\label{fig:supp-quad-showcase}
\end{figure*}

\begin{figure*}[!t]
\centering
\KC{\includegraphics[width=\textwidth]{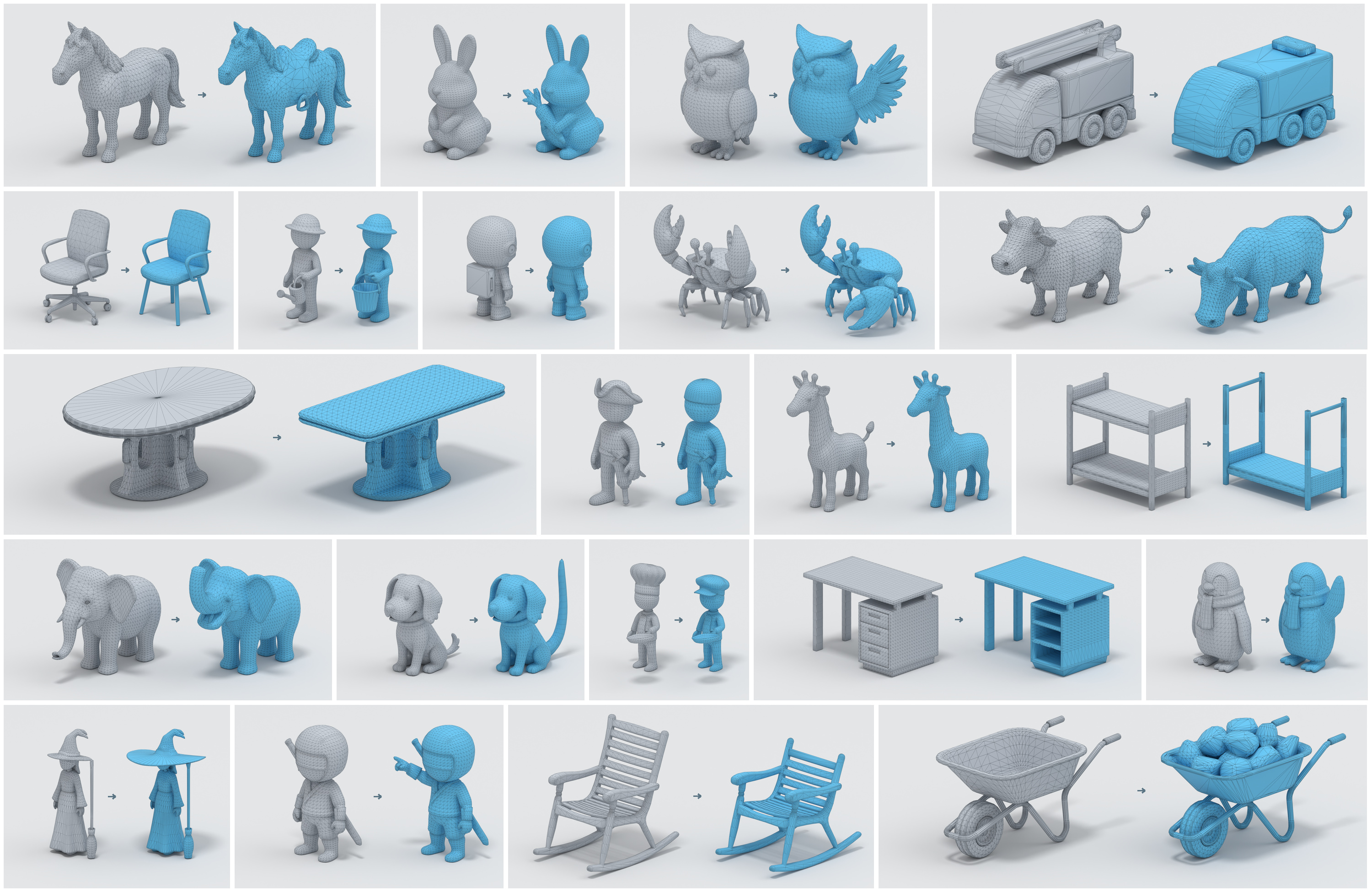}}
\caption{\textbf{Local editing of \name{}-generated meshes.} Each pair shows a generated mesh (left) and its locally edited result (right).}
\label{fig:supp-edit-gallery}
\end{figure*}

\begin{figure*}[!t]
\centering
\includegraphics[width=\textwidth]{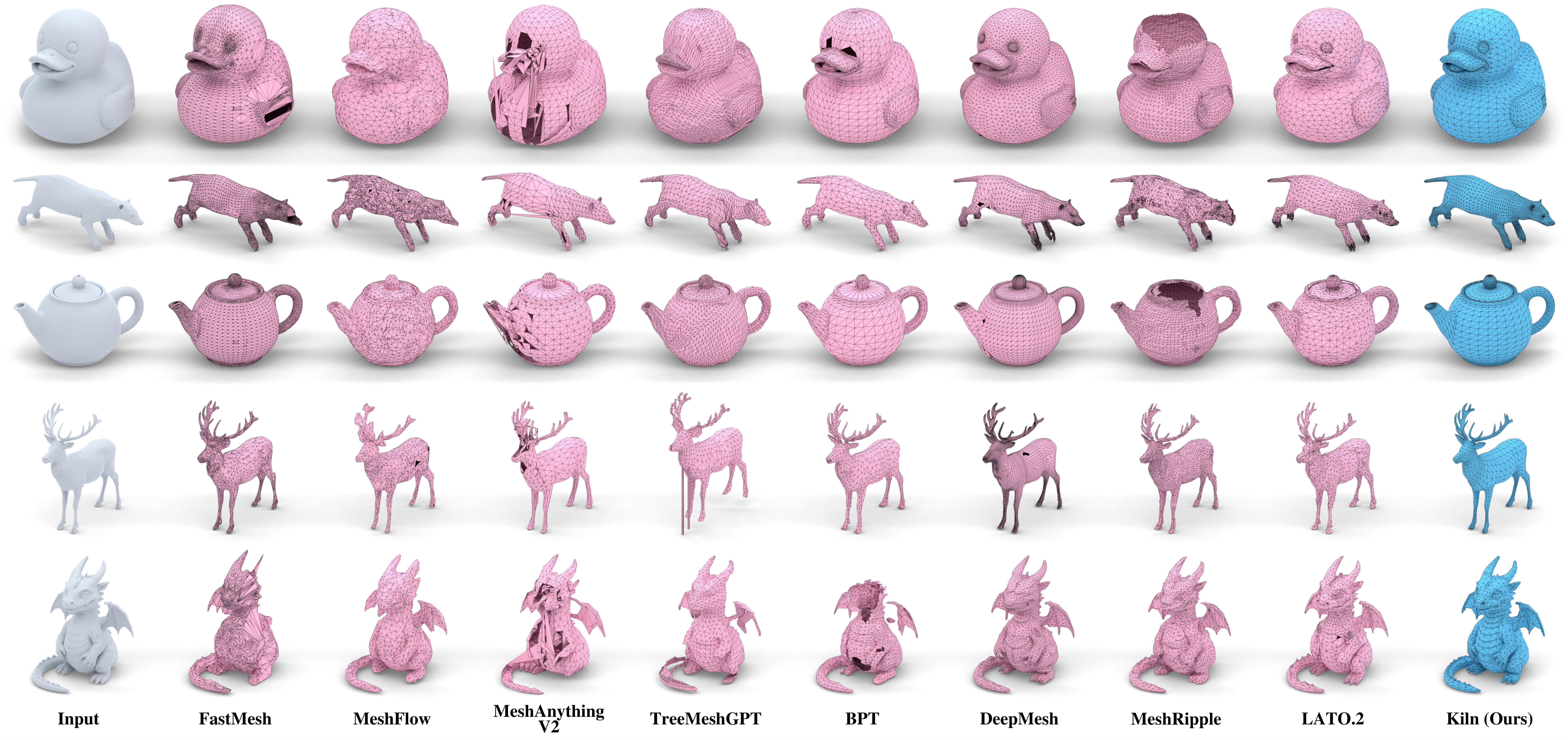}
\caption{\textbf{Additional qualitative comparisons on triangle meshes.}
Our method produces cleaner, better-structured triangle meshes free of holes and non-manifold elements.}
\label{fig:supp_tri_gallery}
\end{figure*}

\begin{figure*}[!t]
\centering
\includegraphics[width=\textwidth]{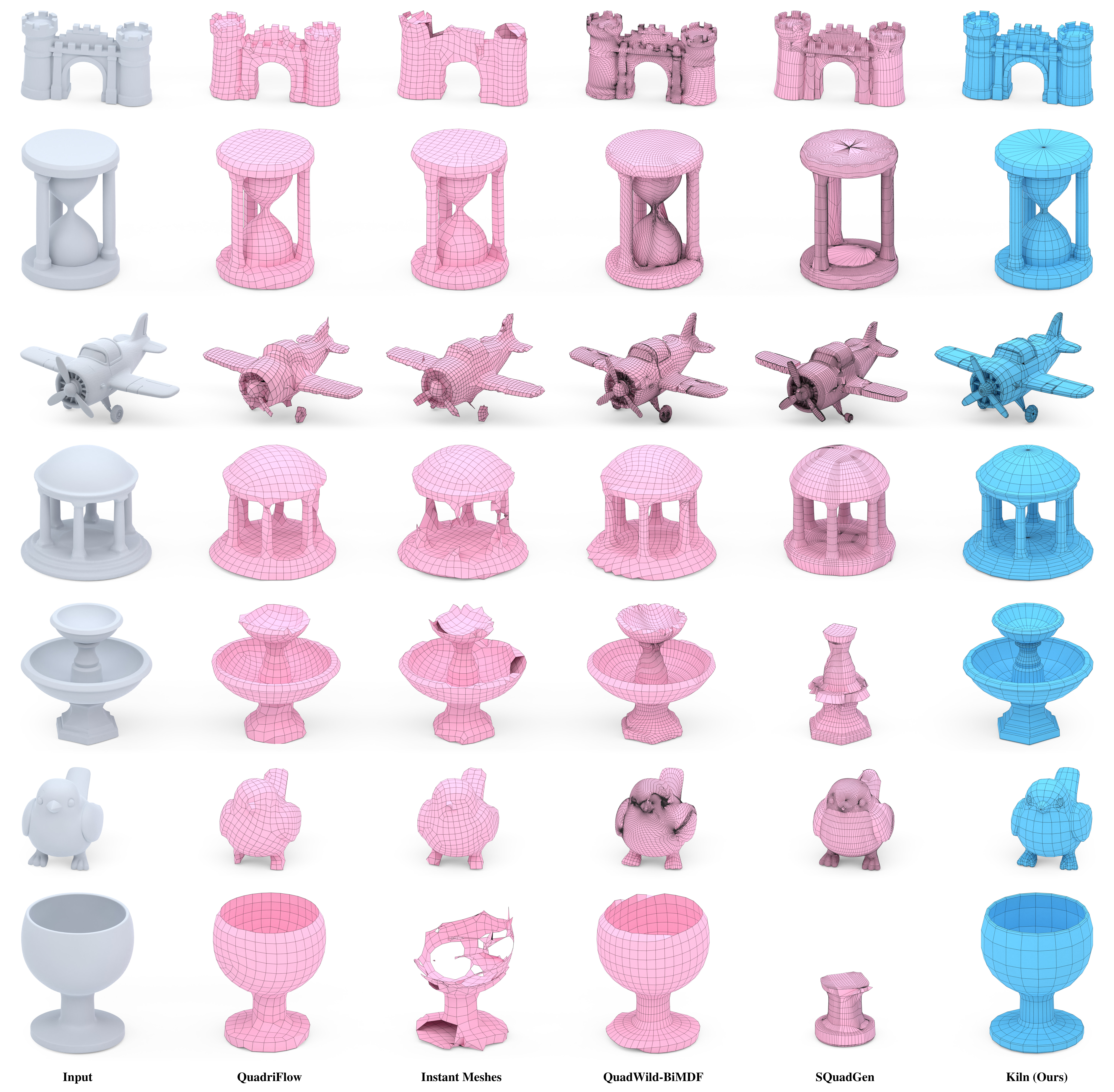}
\caption{\textbf{Additional selected quad-mesh comparison gallery.}
\name{} preserves slender geometric features and produces structured, well-aligned quad layouts.}
\label{fig:supp_quad_gallery}
\end{figure*}

\begin{figure*}[!t]
\centering
\includegraphics[width=0.98\textwidth]{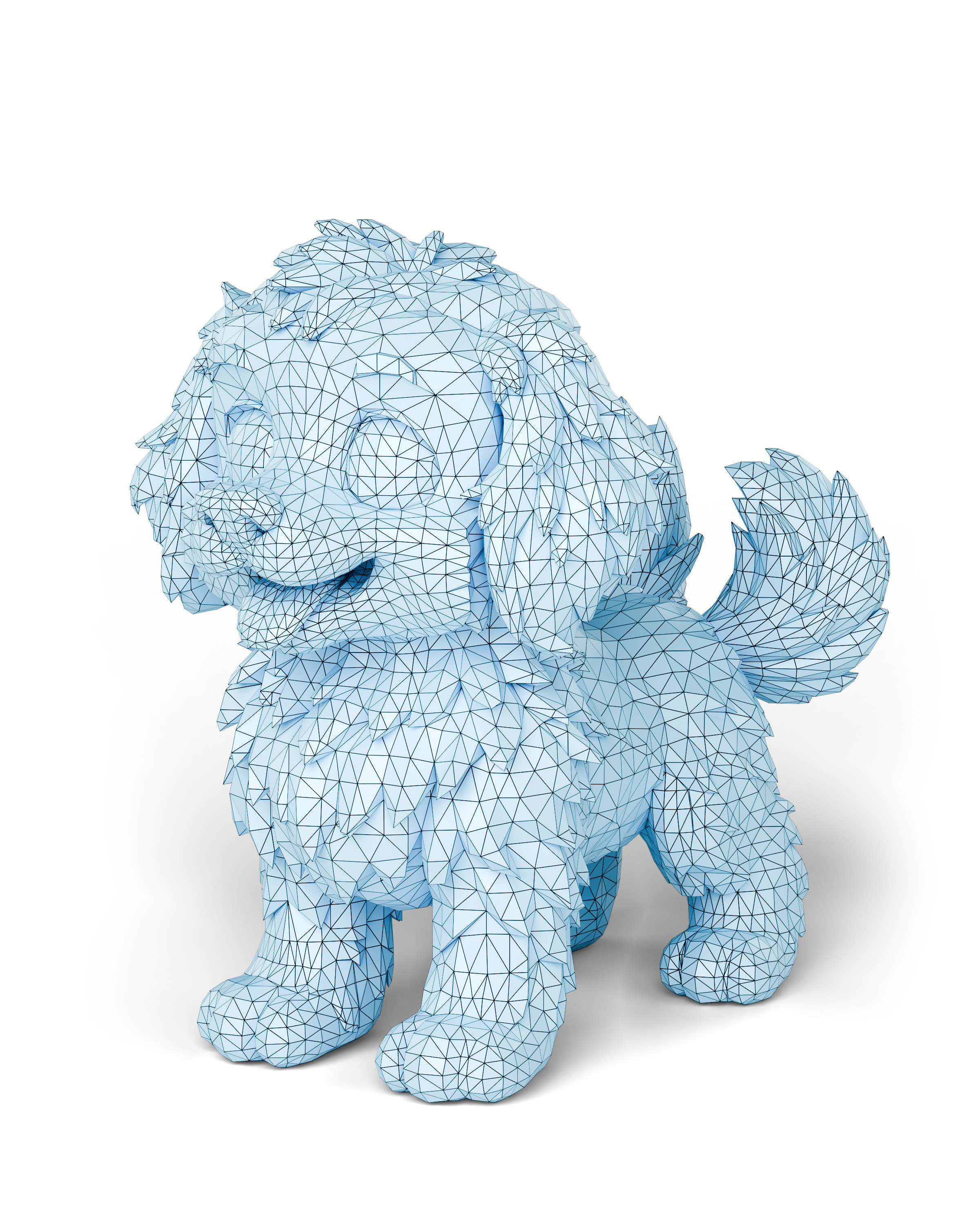}
\caption{\textbf{Additional high-resolution artist mesh showcase.}}
\label{fig:supp_large_1}
\end{figure*}

\begin{figure*}[!t]
\centering
\includegraphics[width=0.98\textwidth]{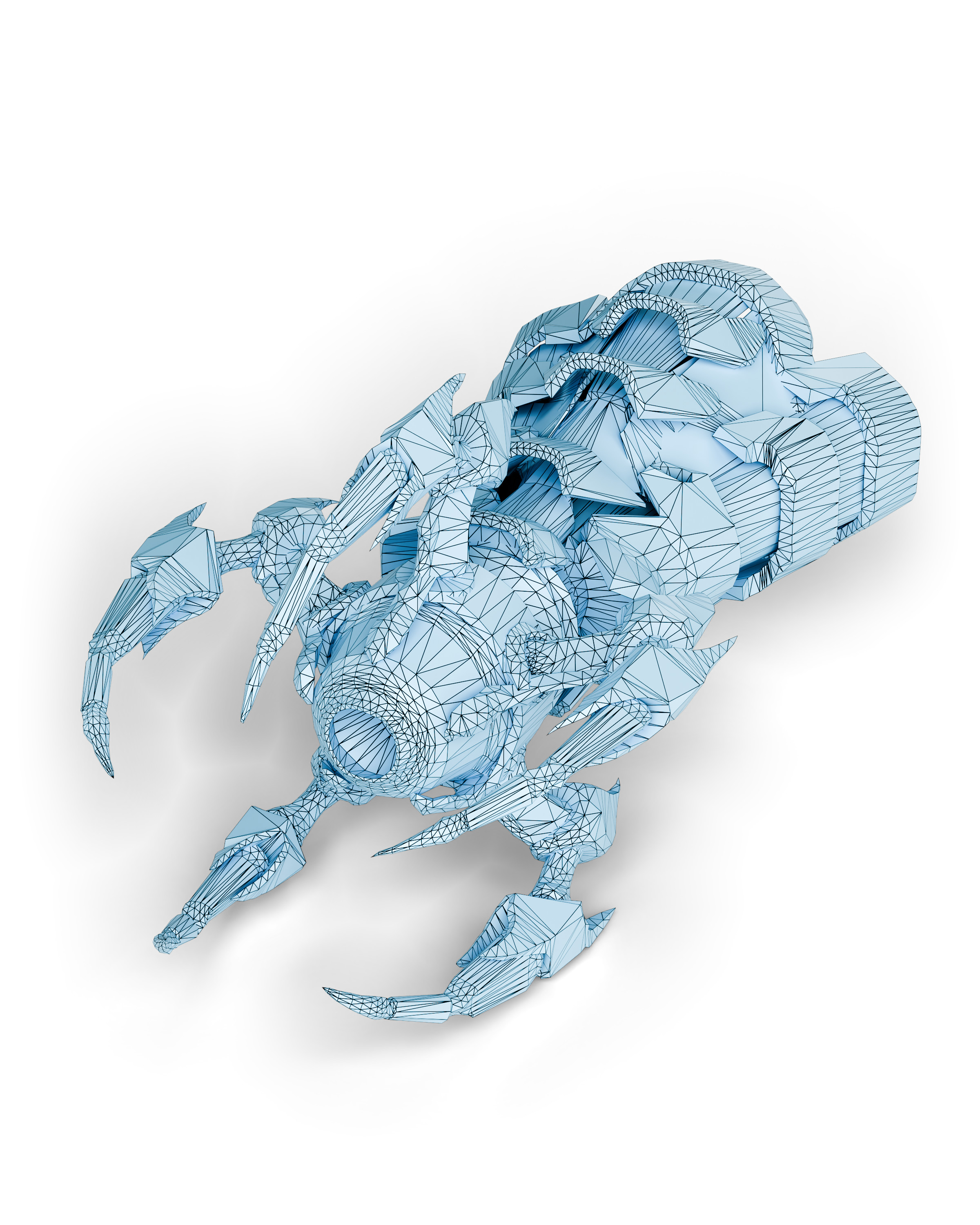}
\caption{\textbf{Additional high-resolution artist mesh showcase.}}
\label{fig:supp_large_2}
\end{figure*}

\begin{figure*}[!t]
\centering
\includegraphics[width=0.98\textwidth]{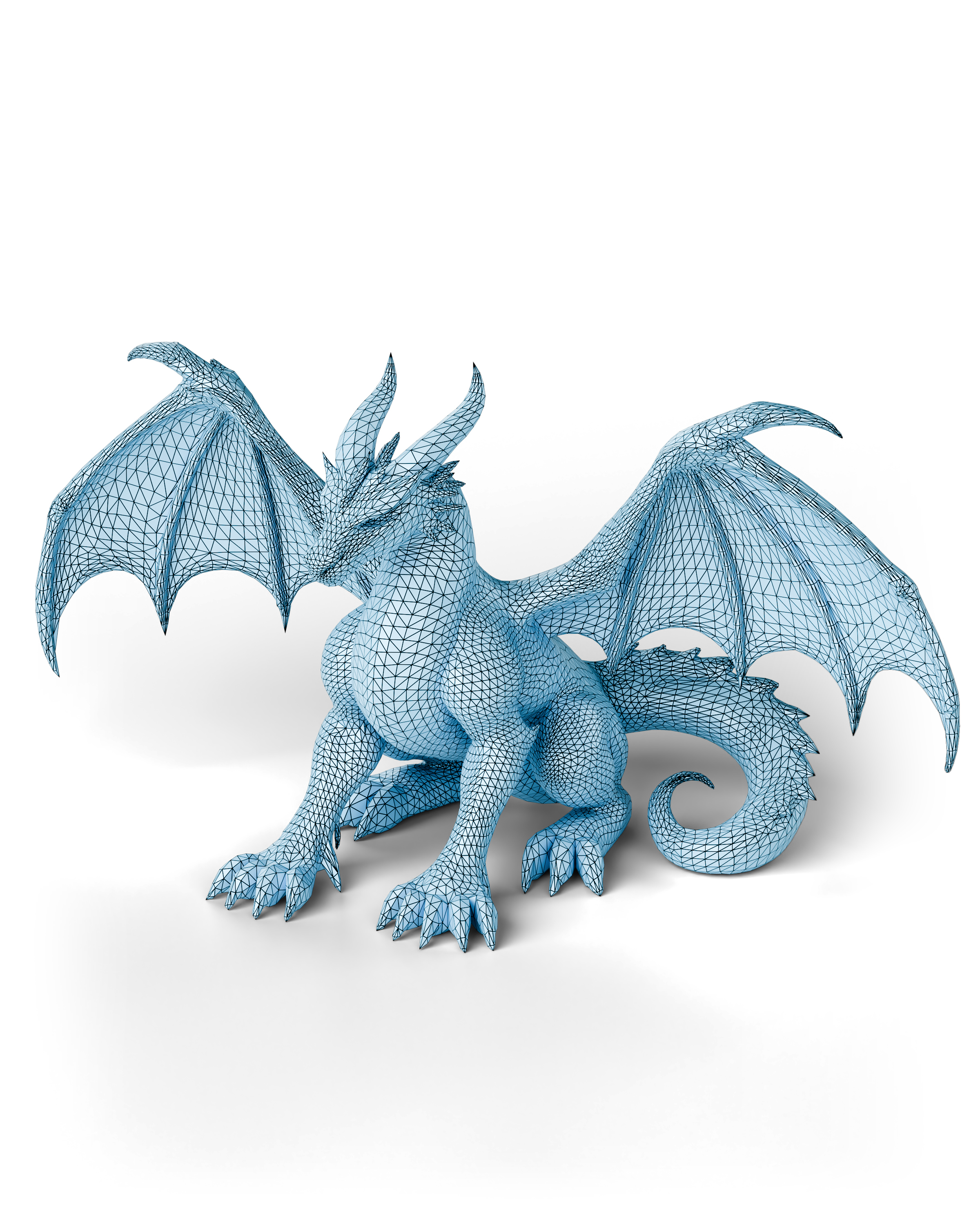}
\caption{\textbf{Additional high-resolution artist mesh showcase.}}
\label{fig:supp_large_3}
\end{figure*}

\begin{figure*}[!t]
\centering
\includegraphics[width=0.98\textwidth]{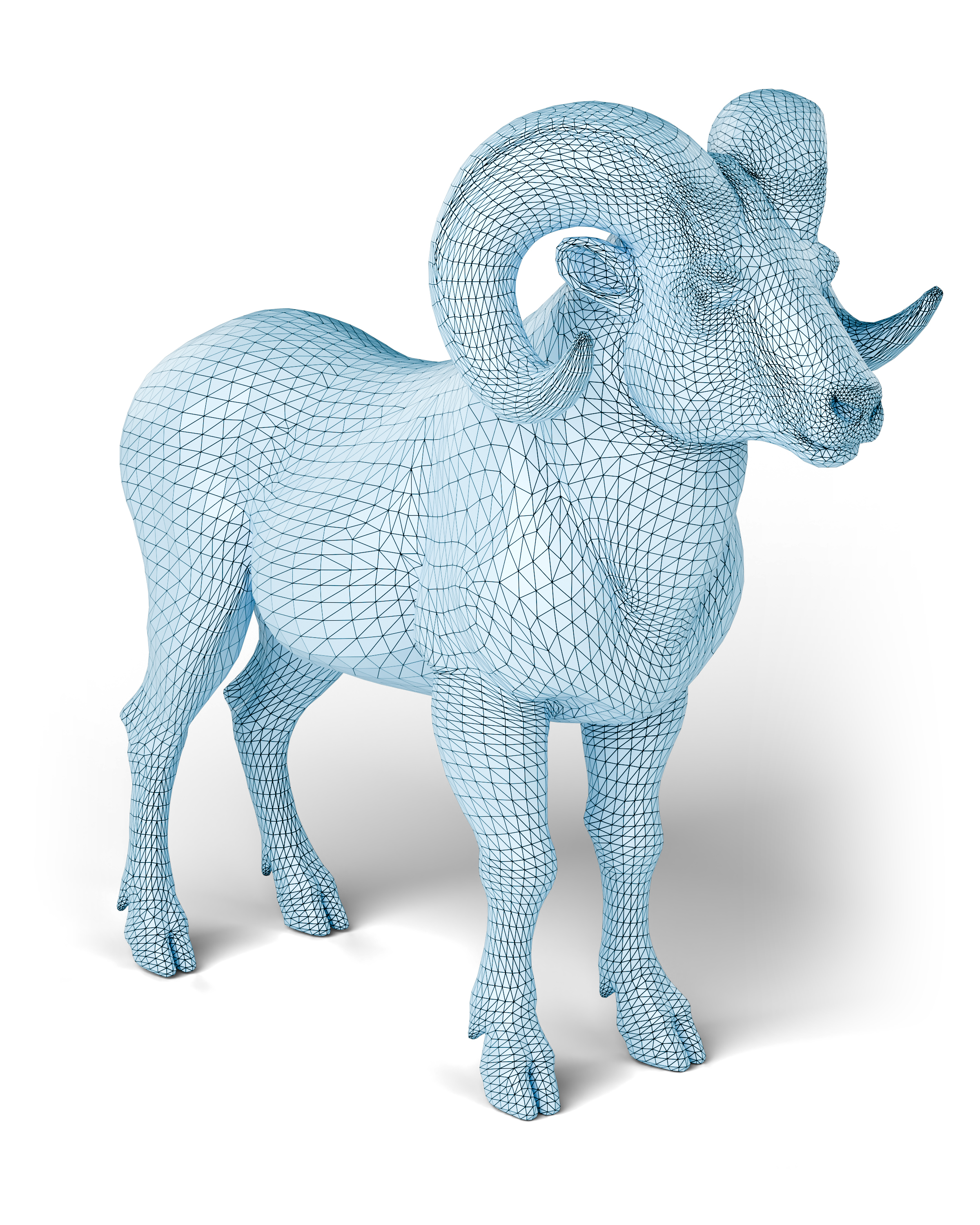}
\caption{\textbf{Additional high-resolution artist mesh showcase.}}
\label{fig:supp_large_4}
\end{figure*}

\begin{figure*}[!t]
\centering
\includegraphics[width=0.98\textwidth]{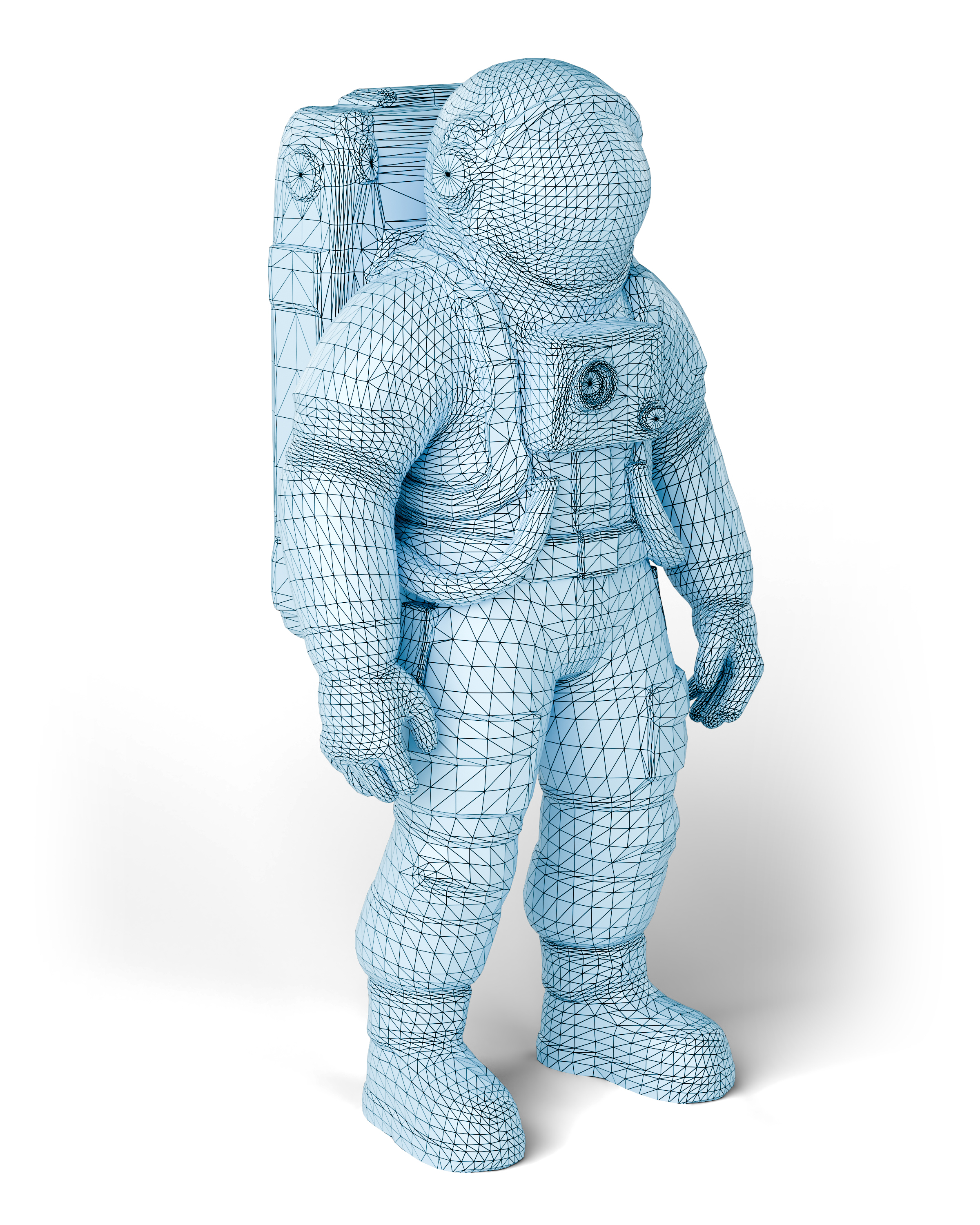}
\caption{\textbf{Additional high-resolution artist mesh showcase.}}
\label{fig:supp_large_5}
\end{figure*}

\Cref{fig:supp-quad-showcase} provides further quad-dominant generation results from \name{}, while \cref{fig:supp-edit-gallery} showcases local shape edits across diverse meshes generated by \name{}, strictly retaining the protected region. In addition, \cref{fig:supp_tri_gallery,fig:supp_quad_gallery} present additional comparison galleries corresponding to \cref{fig:triangle-comparison,fig:quad-comparison} in the main paper.
Finally, \crefrange{fig:supp_large_1}{fig:supp_large_5} highlight a diverse collection of geometrically complex, challenging, and high-face-count meshes synthesized by \name{}.


\clearpage

\clearpage

\clearpage

\clearpage

\clearpage

\end{document}